\documentclass{article}

\usepackage[eandd, final]{neurips_2026}
\usepackage{authblk}
\usepackage{xcolor}
\newsavebox{\FrontierOROneShotTableBox}
\newlength{\FrontierOROneShotNaturalWidth}
\usepackage{colortbl}
\definecolor{FrontierORHighlight}{HTML}{F0EDF9}

\makeatletter
\renewcommand{\@maketitle}{%
  \vbox{%
    \hsize\textwidth
    \linewidth\hsize
    \vskip 0.1in
    \@toptitlebar
    \centering
    {\LARGE\bfseries\@title\par}%
    \@bottomtitlebar
  }%
  {\normalsize \raggedright \@author \par}%
  \vskip 1em%
  \begingroup
    \renewcommand\thefootnote{\ensuremath{\ast}}%
    \footnotetext{Equal contribution.}%
    \renewcommand\thefootnote{\ensuremath{\dagger}}%
    \footnotetext{Project lead.}%
    \renewcommand\thefootnote{\ensuremath{\ddagger}}%
    \footnotetext{Work done outside the authors' employment.}%
    \renewcommand\thefootnote{\ensuremath{\S}}%
    \footnotetext{Joint supervising}%
    \renewcommand\thefootnote{}%
    \footnotetext{Correspondence: \href{mailto:contact@frontieror.org}{contact@frontieror.org}.}%
  \endgroup
  \setcounter{footnote}{0}%
}
\makeatother

\usepackage{tcolorbox}
\tcbuselibrary{skins, breakable}
\usepackage{listings}
\usepackage{enumitem}
\usepackage{booktabs}
\usepackage{makecell}
\usepackage{pifont}
\usepackage[utf8]{inputenc}
\usepackage[T1]{fontenc}
\usepackage{hyperref}
\usepackage{url}
\usepackage{booktabs}
\usepackage{amsfonts}
\usepackage{amsmath}
\usepackage{amssymb}
\usepackage{nicefrac}
\usepackage{microtype}
\usepackage{xcolor}
\usepackage{graphicx}
\usepackage{tabularx}
\usepackage{array}
\usepackage{subcaption}
\usepackage{wrapfig}
\usepackage{float}
\usepackage{needspace}
\usepackage{afterpage}

\newcommand{\CheckmarkBold}{\ding{51}}

\newcommand{\squishlisttwo}[1][$\bullet$]{%
 \begin{list}{#1}{%
    \usecounter{squishlisttwoitem}%
    \setlength{\itemsep}{1.5pt}
    \setlength{\parsep}{1.5pt}
    \setlength{\topsep}{1.5pt}
    \setlength{\parskip}{1.5pt}%
    \setlength{\partopsep}{1.5pt}
    \setlength{\leftmargin}{1.5em}
    \setlength{\labelwidth}{1.5em}
    \setlength{\labelsep}{0.5em}
 }%
}
\newcommand{\squishend}{\end{list}}

\definecolor{taskcardtitle}{HTML}{BEBED4} % source; renders as ~#9E9EB0 (deeper)
\definecolor{taskpanetitle}{HTML}{D0D0E2} % source; renders as ~#AAA9B5 (inner)
\newcounter{casestudy}
\renewcommand{\thecasestudy}{\arabic{casestudy}}
\tcbset{
  taskcard/.style={
    breakable, enhanced, sharp corners,
    colframe=taskcardtitle, boxrule=0.5pt,
    colback=white,
    colbacktitle=taskcardtitle, coltitle=black,
    fonttitle=\bfseries,
    titlerule=0pt,
    title=#1,
    left=4pt, right=4pt, top=4pt, bottom=4pt,
    boxsep=2pt,
  },
}
\tcbset{
  taskpane/.style={
    enhanced, sharp corners,
    colframe=taskpanetitle, boxrule=0.3pt,
    opacityframe=0.7,                          % 30% transparent inner frame
    colback=white,
    colbacktitle=taskpanetitle, coltitle=black,
    fonttitle=\bfseries\footnotesize,
    fontupper=\footnotesize,
    titlerule=0pt,
    left=4pt, right=4pt, top=2pt, bottom=2pt,
    boxsep=1pt,
    title=#1,
  },
}
\tcbset{
  failuremessage/.style={
    enhanced, sharp corners,
    colframe=red!25, colback=red!3,
    colbacktitle=red!12, coltitle=black,
    boxrule=0.3pt, titlerule=0pt,
    fonttitle=\bfseries\scriptsize, fontupper=\scriptsize,
    title={Original message (excerpt)},
    left=3pt, right=3pt, top=2pt, bottom=2pt,
    boxsep=1pt, before skip=2pt, after skip=2pt,
  },
  failurecause/.style={
    enhanced, sharp corners,
    colframe=blue!20, colback=blue!3,
    colbacktitle=blue!10, coltitle=black,
    boxrule=0.3pt, titlerule=0pt,
    fonttitle=\bfseries\scriptsize, fontupper=\scriptsize,
    title={Root cause},
    left=3pt, right=3pt, top=2pt, bottom=2pt,
    boxsep=1pt, before skip=2pt, after skip=0pt,
  },
  codecue/.style={
    failurecause,
    title={Code cue},
  },
}
\lstdefinestyle{taskcode}{
  basicstyle=\ttfamily\scriptsize,
  breaklines=true, breakatwhitespace=true,
  columns=fullflexible, keepspaces=true,
  showstringspaces=false,
  xleftmargin=2pt, xrightmargin=2pt,
  aboveskip=2pt, belowskip=2pt,
  commentstyle=\color{gray!70}\itshape,
  keywordstyle=\color{blue!60!black}\bfseries,
  stringstyle=\color{green!40!black},
  language=Python,
  morekeywords={with,as,True,False,None},
}
\lstdefinestyle{taskjson}{
  basicstyle=\ttfamily\scriptsize,
  breaklines=true, breakatwhitespace=true,
  columns=fullflexible, keepspaces=true,
  showstringspaces=false,
  xleftmargin=2pt, xrightmargin=2pt,
  aboveskip=2pt, belowskip=2pt,
  stringstyle=\color{green!40!black},
  commentstyle=\color{gray!70}\itshape,
}

\hypersetup{
    colorlinks=true,
    linkcolor=blue,
    filecolor=magenta,      
    urlcolor=cyan,
    pdfpagemode=FullScreen,
    citecolor=violet,
}
\setcitestyle{numbers,square,comma}

\title{FrontierOR: Benchmarking LLMs' Capacity for Efficient Algorithm Design in Large-Scale Optimization}

\author[1,2,$\ast$]{Minwei Kong}
\author[1,2,$\ast$]{Chonghe Jiang}
\author[1,2,$\dagger$]{Ao Qu}
\author[2]{Wenbin Ouyang}
\author[3]{Zhaoming Zeng}
\author[4,$\ddagger$]{Xiaotong Guo}
\author[2]{Zhekai Li}
\author[1]{Junyi Li}
\author[5]{Yi Fan}
\author[6]{Xinshou Zheng}
\author[6]{Xi Jing}
\author[6]{Yikai Zhang}
\author[7]{Zhiwei Liang}
\author[8]{Seonghoo Kim}
\author[6]{Runqing Yang}
\author[9]{Zijian Zhou}
\author[10,$\ddagger$]{Sirui Li}
\author[2]{Han Zheng}
\author[11]{Wangyang Ying}
\author[11]{Ou Zheng}
\author[12]{Chonghuan Wang}
\author[6]{Jinglong Zhao}
\author[9]{Hanzhang Qin}
\author[2]{Cathy Wu}
\author[1,2]{Paul Pu Liang}
\author[1,2,$\S$]{Jinhua Zhao}
\author[1,13,$\S$]{Hai Wang}

\affil[1]{Singapore-MIT Alliance for Research and Technology}
\affil[2]{Massachusetts Institute of Technology}
\affil[3]{Northeastern University}
\affil[4]{Uber}
\affil[5]{Shanghai Jiaotong University}
\affil[6]{Boston University}
\affil[7]{Emory University}
\affil[8]{Northwestern University}
\affil[9]{National University of Singapore}
\affil[10]{Microsoft}
\affil[11]{Zhiling Research}
\affil[12]{University of Texas at Dallas}
\affil[13]{Singapore Management University}

\begin{document}

\maketitle

\begin{abstract}
Large language models (LLMs) are increasingly used for optimization modeling and solver-code generation, yet practical operations research and optimization problems often require a harder capability: designing scalable algorithms that exploit problem structure and outperform direct formulation-and-solve baselines. Existing benchmarks are limited to small or simplified examples far below real-world scale and complexity. We introduce \textbf{FrontierOR}, the first literature-grounded benchmark targeting LLM-generated algorithm efficiency on realistic large-scale optimization problems. FrontierOR includes 180 diverse tasks, each derived from a paper published in a top-tier operations research venue, with standardized instances and a hidden, expert-verified evaluation suite. We evaluate recent LLMs' one-shot algorithm generation and improvements achieved through several self-evolving frameworks. The results show that current models still struggle to move from executable formulations to efficient optimization algorithms. Even the strongest LLM succeeds on only 39\% of Full-set task instances, where success is defined as returning a feasible solution with an objective gap of at most 1\% relative to Gurobi within its runtime. On the Hard subset, test-time evolution raises this proportion from 51\% for the strongest one-shot model to 78\%. FrontierOR provides a continuously maintained testbed for evaluating LLM-designed optimization algorithms under a shared protocol and controlled computational settings. Code and data are publicly released at \url{https://github.com/Minw913/FrontierOR}.

\end{abstract}

\begin{figure}[!h]
\centering
\includegraphics[width=\linewidth]{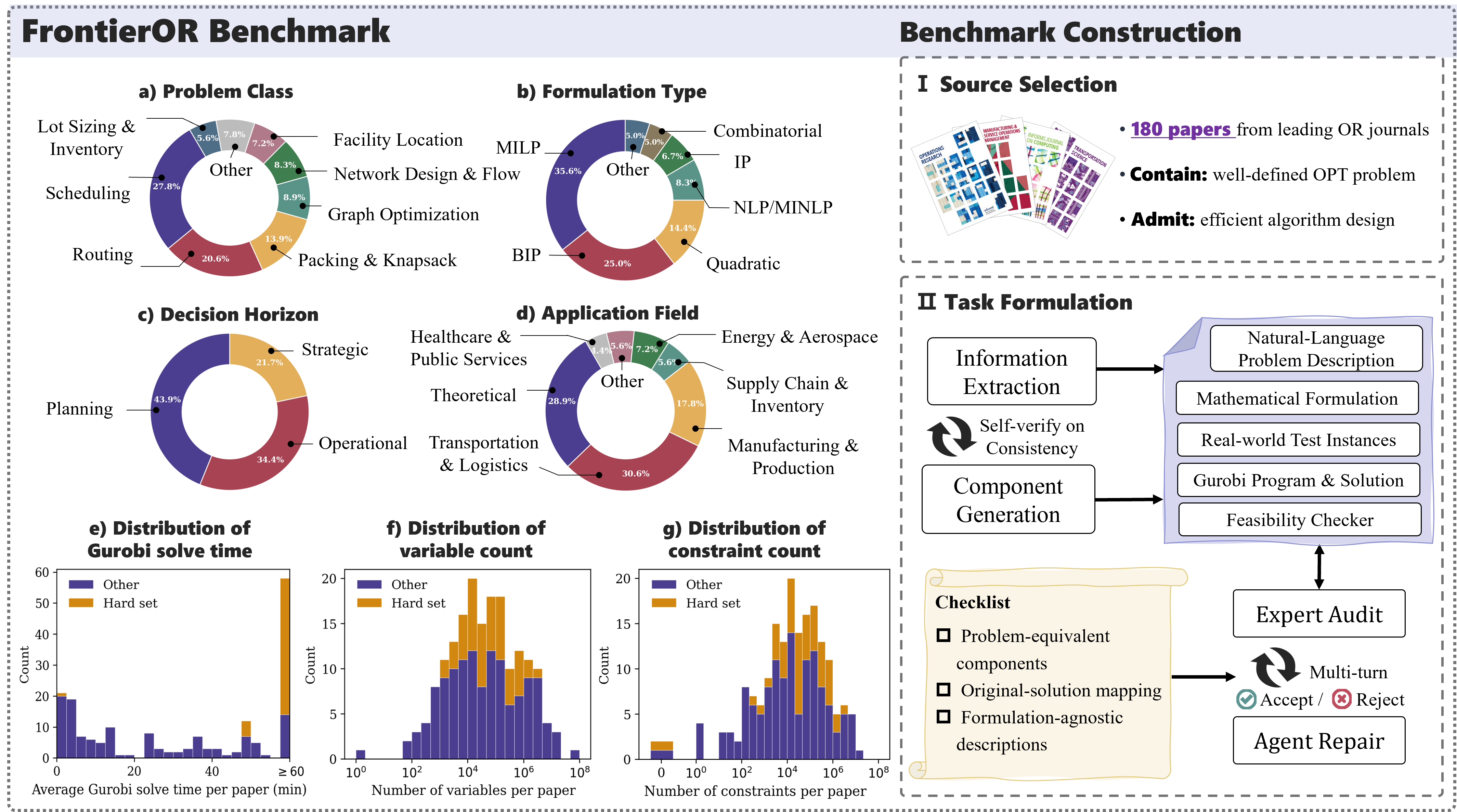}
\caption{Overview of the FrontierOR benchmark. FrontierOR spans diverse problem domains, formulation types, and application fields. In panels~(e), (f), and~(g), the vertical axis \emph{Count} denotes the number of FrontierOR tasks. Those optimization tasks contain up to $10^8$ variables and over $10^7$ constraints. $48.1\%$ of the large-instance reference runs exceed the one-hour runtime budget. We construct the benchmark by collecting problems from leading OR journals, and ensure data quality through multi-round expert review.}

\label{fig:frontieror-overview}
\end{figure}

\section{Introduction}
Recent progress in large language models (LLMs) has stimulated growing interest in using LLMs for operations research (OR). Most existing work studies LLMs as modeling assistants, evaluating whether they can translate natural-language decision problems into mathematical optimization formulations or solver-ready programs \citep{ramamonjison2023nl4opt,ahmaditeshnizi2024optimus,yang2024optibench,michailidis2025cp,kong2025alphaopt,huang2025orlm,zhou2025steporlm,xiao2023chain,jiang2024llmopt}. 
Yet practical OR also requires algorithms that exploit problem structure and scale under realistic computational constraints, leaving a central capability largely under-measured: \textit{Can LLMs read a real-world decision problem description and implement an efficient algorithm?}
Answering this question requires a systematic evaluation framework with \textit{realistic} problem description, \textit{diverse} problem coverage, \textit{scalable} instances, and \textit{rigorous} evaluators.
However, existing benchmarks address these requirements only in part: many focus on direct solution reasoning \citep{tang2024grapharena,jiang2026reasoning} or formulation correctness \citep{li2026industrial} rather than algorithmic efficiency. Those explicitly targeting algorithm design remain limited to relatively compact collections of optimization problems \citep{sun2026co,chen2025heurigym}.

To this end, we introduce \textbf{FrontierOR}, a benchmark for evaluating optimization algorithm efficiency on realistic large-scale OR problems (Figure~\ref{fig:frontieror-overview}).
Unlike previous works that primarily evaluate formulation accuracy on textbook-scale modeling tasks, FrontierOR contains 180 tasks derived from top-tier OR literature, with instances spanning up to ${\sim}10^8$ decision variables and ${\sim}10^7$ constraints.
The tasks are selected because they contain exploitable algorithmic structure---for example, decomposability, temporal or network coupling, sparsity, routing and assignment structure, or scenario structure---so solving them effectively requires more than constructing a monolithic solver formulation.
FrontierOR therefore evaluates generated solver programs not only by feasibility, but also by solution quality and runtime competitiveness against an expert-verified Gurobi baseline, directly targeting algorithm efficiency rather than formulation accuracy alone.
To further ensure benchmark quality, after information extraction from the source papers, each task is annotated and verified by 15 PhD researchers in operations research through a three-week, multi-turn review process in which each task is assigned to a primary reviewer with matched subfield expertise.

Each FrontierOR task pairs a model-facing specification with a hidden evaluation suite. The model receives only a natural-language problem description and input/output formats; the mathematical formulation, test instance data, the reference solver implementation, and algorithmic guidance are withheld during code generation. 
Evaluation proceeds in two stages. First, we assess \textit{correctness} using the task-specific feasibility checker to verify that the generated program's returned solution satisfies variable-domain restrictions and all hard constraints. Second, for feasible solutions, we assess \textit{algorithmic performance} along two dimensions: solution quality (objective gap relative to the Gurobi reference solution) and computational efficiency (runtime relative to Gurobi under the same resource budget).
Our main quality--time efficiency (QTE) metric counts an instance as successful only when the generated algorithm returns a feasible solution with an objective gap of at most 1\% relative to the Gurobi reference, at least as fast as the reference.

We evaluate twelve recent LLMs under one-shot generation and test-time evolution. In the one-shot setting, the model generates a solver program from the problem description alone. In the self-evolve setting, an agent iteratively revises candidate programs using development-set feedback before final scoring on held-out instances. The results show that current models remain far from reliable algorithm designers: the strongest one-shot model achieves QTE on only 39\% of Full-set task instances. Stronger models more often generate decomposition, local-search, or hybrid algorithms, whereas weaker models tend to rely on monolithic solver calls. On the Hard subset, the best self-evolving framework improves feasibility, solution quality, and QTE over the one-shot program by 27--31 percentage points. These gains demonstrate the value of feedback-driven test-time evolution, while cross-instance generalization leaves room for improvement.

In summary, we introduce a benchmark that evaluates LLMs on large-scale optimization algorithm design, shifting the focus from modeling correctness alone to feasible, scalable, and efficient algorithm generation. Our main contributions are listed as follows:
\squishlisttwo[\arabic*.]
\item We construct and publicly release a broad, literature-grounded benchmark, FrontierOR, which consists of 180 real-world OR tasks spanning diverse industrial problem settings, each with natural-language specifications, large-scale instances, math formulations, verified Gurobi baselines, and standalone feasibility checkers.
\item We design a systematic evaluation pipeline for FrontierOR and conduct extensive experiments to assess the capabilities of widely used LLMs as well as three representative test-time evolution methods. We perform a comprehensive analysis using multi-dimensional metrics, including feasibility, solution quality, and quality–time efficiency.
\squishend

\section{Related Work} 
\label{sec:related-work}
\textbf{LLMs for optimization modeling and solving.}
Benchmarks such as NL4OPT, OptiBench, MAMO, and CP-Bench evaluate the translation of natural-language problems into mathematical models or solver programs \citep{ramamonjison2023nl4opt,yang2024optibench,huang2025llms,michailidis2025cp}. Methods improve modeling through agentic debugging (OptiMUS), training and data synthesis (ORLM and OptMATH), experience reuse (AlphaOPT), structured workflows (LEAN-LLM-OPT), and process supervision (StepORLM) \citep{ahmaditeshnizi2024optimus,huang2025orlm,lu2025optmath,kong2025alphaopt,liang2026large,zhou2025steporlm}. FrontierOR extends the focus from model construction to scalable, efficient algorithm design.

\textbf{LLMs for algorithm design and heuristic search.}
FunSearch and AlphaEvolve combine code generation, execution-based evaluation, and evolutionary search \citep{romera2024mathematical,novikov2025alphaevolve}. EoH jointly evolves algorithmic ideas and code, ReEvo adds reflective feedback, and MCTS-AHD explores heuristics through tree search \citep{liu2024evolution,ye2024reevo,zheng2025monte}. Complementing these methods, FrontierOR evaluates end-to-end algorithm design from natural-language descriptions, allowing autonomous selection among solution strategies (e.g., exact algorithms, heuristics, decomposition, and hybrid methods).

\textbf{Combinatorial optimization benchmark suites.}
Benchmark suites, which enable reproducible solver comparisons and algorithm testing beyond hand-picked instances, have long been central to empirical research in combinatorial optimization. Classical resources such as OR-Library~\citep{beasley1990or}, TSPLIB~\citep{reinelt1991tsplib}, CVRPLIB~\citep{uchoa2017new}, QAPLIB~\citep{burkard1997qaplib}, and MIPLIB~\citep{gleixner2021miplib} remain indispensable for evaluating exact and heuristic algorithms. However, these resources primarily diversify instances within established problem classes rather than broaden problem coverage, and were not designed specifically to evaluate LLMs.

\textbf{LLM benchmarks for optimization algorithm design.}
Recent benchmarks evaluate LLMs and learning-based solvers across graph and combinatorial reasoning, optimization modeling, solver performance, and algorithm generation \citep{tang2024grapharena,jiang2026reasoning,li2026industrial,feng2025comprehensive,sun2026co,chen2025heurigym}. Those explicitly targeting algorithm design provide valuable evaluations but cover relatively compact task collections. FrontierOR complements these efforts by evaluating scalable algorithm design across 180 diverse, literature-grounded OR tasks, jointly assessing feasibility, solution quality, and solving time. Table~\ref{tab:benchmark-comparison} provides a detailed comparison.

\begin{table}[h]
\centering
\vspace{-0.25cm}
\caption{Comparison with representative LLM-for-optimization benchmarks.}
\label{tab:benchmark-comparison}
\setlength{\tabcolsep}{4pt}
\renewcommand{\arraystretch}{0.95}
\small
\resizebox{\textwidth}{!}{%
\begin{tabular}{lccccccc}
\toprule
\textbf{Benchmark}
& \textbf{End-to-End}
& \textbf{\#Problems}
& \textbf{Lit.-Grounded}
& \textbf{NL Input}
& \textbf{Large-Scale}
& \textbf{Solver Baseline}
& \textbf{Eval.\ Protocol} \\
\midrule

GraphArena~\citep{tang2024grapharena}
& & 10 & & \CheckmarkBold & & & \\

CO-Bench~\citep{sun2026co}
& \CheckmarkBold & 36 & & \CheckmarkBold & & \CheckmarkBold & \CheckmarkBold \\

FrontierCO~\citep{feng2025comprehensive}
& & 8 & & & \CheckmarkBold & \CheckmarkBold & \CheckmarkBold \\

HeuriGym~\citep{chen2025heurigym}
& \CheckmarkBold & 9 & & \CheckmarkBold & & & \\

NLCO~\citep{jiang2026reasoning}
& & 43 & & \CheckmarkBold & & \CheckmarkBold & \\

MIPLIB-NL~\citep{li2026industrial}
& \CheckmarkBold & 223 & & \CheckmarkBold & \CheckmarkBold & \CheckmarkBold & \\

\midrule
\textbf{FrontierOR (Ours)}
& \CheckmarkBold & \textbf{180} & \CheckmarkBold & \CheckmarkBold & \CheckmarkBold & \CheckmarkBold & \CheckmarkBold \\
\bottomrule
\end{tabular}%
}
\end{table}

\begin{minipage}{\textwidth}
\footnotesize
\textit{Note.}
\textbf{End-to-End}: the LLM writes solution code from scratch given the input.
\textbf{\#Problems}: number of optimization problems.
\textbf{Lit.-Grounded}: benchmark tasks are extracted or adapted from peer-reviewed papers.
\textbf{NL Input}: the LLM receives a natural-language problem description.
\textbf{Large-Scale}: the test instances are large-scale.
\textbf{Solver Baseline}: the benchmark provides reference solver results for comparison.
\textbf{Eval.\ Protocol}: the benchmark provides a complete evaluation protocol consisting of a feasibility checker and metrics for measuring solution quality and efficiency.
\end{minipage}
\section{FrontierOR Benchmark}
\label{sec:benchmark}
This section describes the construction pipeline of FrontierOR. 
The dataset covers diverse problem classes, application domains, and instance scales. The task-formulation pipeline converts selected OR papers into algorithm-design tasks through information extraction, component generation, and human validation, producing standardized instances, Gurobi references, and feasibility checkers. Additional construction details are provided in Appendix~\ref{app:construction-details}.

\subsection{Dataset Description}
\label{subsec:collection}

\textbf{Source selection criteria.}
We curate tasks from leading OR journals, including \emph{Operations Research}, \emph{Management Science}, \emph{Transportation Science}, \emph{INFORMS Journal on Computing}, \emph{European Journal of Operational Research}, alongside other publication venues.
% ao: add a number. how many other related journals?
As shown in Figure~\ref{fig:frontieror-overview}, each selected paper studies a well-defined optimization problem, provides sufficient information for task construction, and motivates efficient algorithm design beyond directly applying a generic solver. These criteria ensure that every task is grounded in a real optimization setting in which scalable algorithm design is practically important.
We further assess potential source-paper memorization through a paper-memory audit and a common-subset analysis excluding potentially memorized tasks (Appendix~\ref{app:contamination}).

\textbf{Dataset statistics.}
\label{subsec:statistics}
FrontierOR contains 180 literature-grounded tasks from 1992--2025, including 69 papers published in 2020 or later. Figure~\ref{fig:frontieror-overview} demonstrates that the collected benchmark tasks span diverse problem classes (e.g., routing, scheduling, and facility location), primary formulation families (e.g., MILP, BIP, and quadratic programming), and a wide range of application fields (e.g., transportation, energy, supply chain, and healthcare). Each task includes one tiny instance and five large instances. The large instances are computationally hard: a median problem has roughly $40{,}000$ decision variables and $18{,}000$ constraints on its large instance, and under a one-hour evaluation budget, $52.3\%$ ($471/900$) of the large-instance references are recorded as feasible incumbents rather than certified optima. Among the 471 incumbent references, the median available MIP gap is $24.3\%$.
Each task is also annotated as planning,
strategic, or operational to characterize
its typical decision-making timescale and deployment context,
reflecting diverse priorities for solution quality and computational efficiency.
Additional statistics are relegated in Appendix~\ref{app:dataset-statistics}.

\subsection{Task Formulation}
\label{subsec:task}
We then describe how FrontierOR tasks are constructed from source papers through a pipeline of automated component generation followed by expert quality assurance.

\textbf{Component generation.}
As shown in Figure~\ref{fig:frontieror-overview}, each selected paper is converted into a suite of task components through information extraction and component generation by Claude Code (Opus 4.6). We extract the natural-language problem description and mathematical formulation, then use the formulation as the primary reference for constructing a standardized test-instance format, a Gurobi implementation with reference solutions, and a standalone feasibility checker, see examples in Appendix~\ref{app:dataset-example}. The problem description is synthesized from the source paper, mathematical formulation, and input/output schemas, but avoids explicit mathematical notation and formulation-specific modeling terminology.
We assess potential Claude-specific wording bias by comparing original and Codex-regenerated problem descriptions as inputs to the evaluated LLMs (Appendix~\ref{app:construction-bias}).

\textbf{Quality assurance.}
We apply two quality-assurance layers. First, automatic verification cross-runs each feasibility checker against all Gurobi reference solutions, with any inconsistency triggering repair over the checker, Gurobi code, data parsing, numerical tolerance, and extracted formulation. Second, 15 OR experts audit formulation fidelity, description completeness, and Gurobi-code alignment against the source paper, and verify that each checker enforces the original problem constraints. After iterative repair and re-verification, all retained tasks have Gurobi solutions that pass their checkers and LLM-facing descriptions sufficient for executable algorithm design. The full checklist and common error categories are reported in Appendix~\ref{app:qa-details}.

\textbf{Full/ Hard partition.}
We curate a \emph{Hard} subset of 50 tasks from FrontierOR to evaluate algorithm performance on genuinely difficult OR problems. Tasks are selected using three criteria:
\emph{(1) Problem class with combinatorial blow-up.} Classical NP-hard families that resist monolithic MIP at realistic scales (e.g., lot-sizing, graph optimization, and sequencing scheduling).
\emph{(2) Instance scale.} Tasks whose composite scale score, computed from the percentile ranks of variable, integer-variable, and constraint counts, is at or above the median across all tasks.
\emph{(3) Empirical solver saturation.} Historical Gurobi records indicate non-optimal termination within a 1-hour time budget or an upper-quartile median runtime.
Tasks satisfying either structural criterion (1) or (2), together with (3), form the hard-candidate pool. We rank these candidates using a composite score combining solver saturation, instance scale, and problem structure, and select the top 50.

\section{Experimental Setup}
\label{sec:experimental-setup}
\subsection{Computational Environment}
\label{subsec:computational-environment}
To ensure that optimization programs are compared on algorithmic merit rather than implementation-level parallelism, we adopt a Docker-based execution environment that enforces strict, reproducible resource constraints. Each optimization program evaluation run is allocated a single CPU core and a one-hour execution time limit within a fixed container image shipping Python~3.13 and Gurobi~12 as the sole optimization solver, together with standard scientific libraries (e.g., NumPy and SciPy) for data manipulation. Network access is disabled to prevent generated code from reaching external resources during evaluation.
For both Gurobi reference runs and Gurobi calls within LLM-generated programs, we enforce the same solver settings: \texttt{Seed=0} for reproducible randomized behavior, \texttt{MIPGap=1e-4} for the relative optimality-gap tolerance, \texttt{Threads=1} for single-threaded execution, and \texttt{NumericFocus=0} for Gurobi's default balance between numerical caution and solve speed.

\subsection{Evaluation Metrics}
\label{subsec:eval-metrics}
We evaluate LLM-generated algorithms using the four metrics summarized in Table~\ref{tab:evaluation-metrics}. Let $N$ denote the number of tasks and $M$ the total number of large instances across all tasks. For task $k$, $E_k$ indicates successful program execution on the tiny instance. For large instance $i$, $F_i$ indicates whether the program returns a feasible solution within the time limit; $f_i$ and $f_i^{\mathrm{G}}$ denote the objective values returned by the generated program and Gurobi reference, respectively; and $t_i$ and $t_i^{\mathrm{G}}$ denote their wall-clock runtimes.

We define the signed relative solution gap for each feasible case as
\begin{equation}
g_i = d_i\frac{f_i-f_i^{\mathrm{G}}}{\max\!\left(\lvert f_i^{\mathrm{G}}\rvert,\epsilon\right)},
\qquad
d_i =
\begin{cases}
+1, & \text{for minimization},\\
-1, & \text{for maximization}.
\end{cases}
\label{eq:solution-gap}
\end{equation}
where $\epsilon>0$ prevents division by zero. Thus, $g_i>0$ means that the generated solution is worse than the Gurobi reference, whereas $g_i\leq 0$ means that it matches or improves upon the reference. We use $g_i\leq0.01$ (a $1\%$ gap) as the solution-quality threshold, with a sensitivity analysis of this choice provided in Appendix~\ref{app:gap-threshold-sensitivity}.

\begin{table}[h]
\centering
\caption{Summary of the evaluation metrics. All metrics are reported as rates in $[0,1]$, with higher values indicating better performance.}
\label{tab:evaluation-metrics}
\small
\setlength{\tabcolsep}{5pt}
\begin{tabularx}{\textwidth}{l X c}
\toprule
\textbf{Metric} & \textbf{Definition} & \textbf{Aggregation level} \\
\midrule
Execution rate & $N^{-1}\sum_{k=1}^{N}E_k$: fraction of tasks whose generated program executes without a runtime error on the tiny instance after up to five runtime-error self-debugging iterations. & Task \\
Feasibility & $M^{-1}\sum_{i=1}^{M}F_i$: fraction of large-instance cases returning a feasible solution within the time limit. & Large instance \\
Solution quality & $M^{-1}\sum_{i=1}^{M}F_i\mathbb{I}[g_i\leq0.01]$: fraction of large-instance cases returning feasible solutions with an objective gap of at most $1\%$ relative to the Gurobi reference. & Large instance \\
QTE & $M^{-1}\sum_{i=1}^{M}F_i\mathbb{I}[g_i\leq0.01]\mathbb{I}[t_i\leq t_i^{\mathrm{G}}]$: fraction of large-instance cases in which the program returns a feasible solution with an objective gap of at most $1\%$ relative to the Gurobi reference, at least as fast as the reference. & Large instance \\
\bottomrule
\end{tabularx}
\end{table}

Here, $\mathbb{I}[\cdot]$ is the indicator function. Infeasible, missing, or timed-out cases have $F_i=0$ and receive no solution-quality or QTE credit.

\subsection{Benchmarked Models and Evaluation Protocols}
\label{subsec:eval-protocol}
\begin{figure}[!h]
\centering
\includegraphics[width=\textwidth]{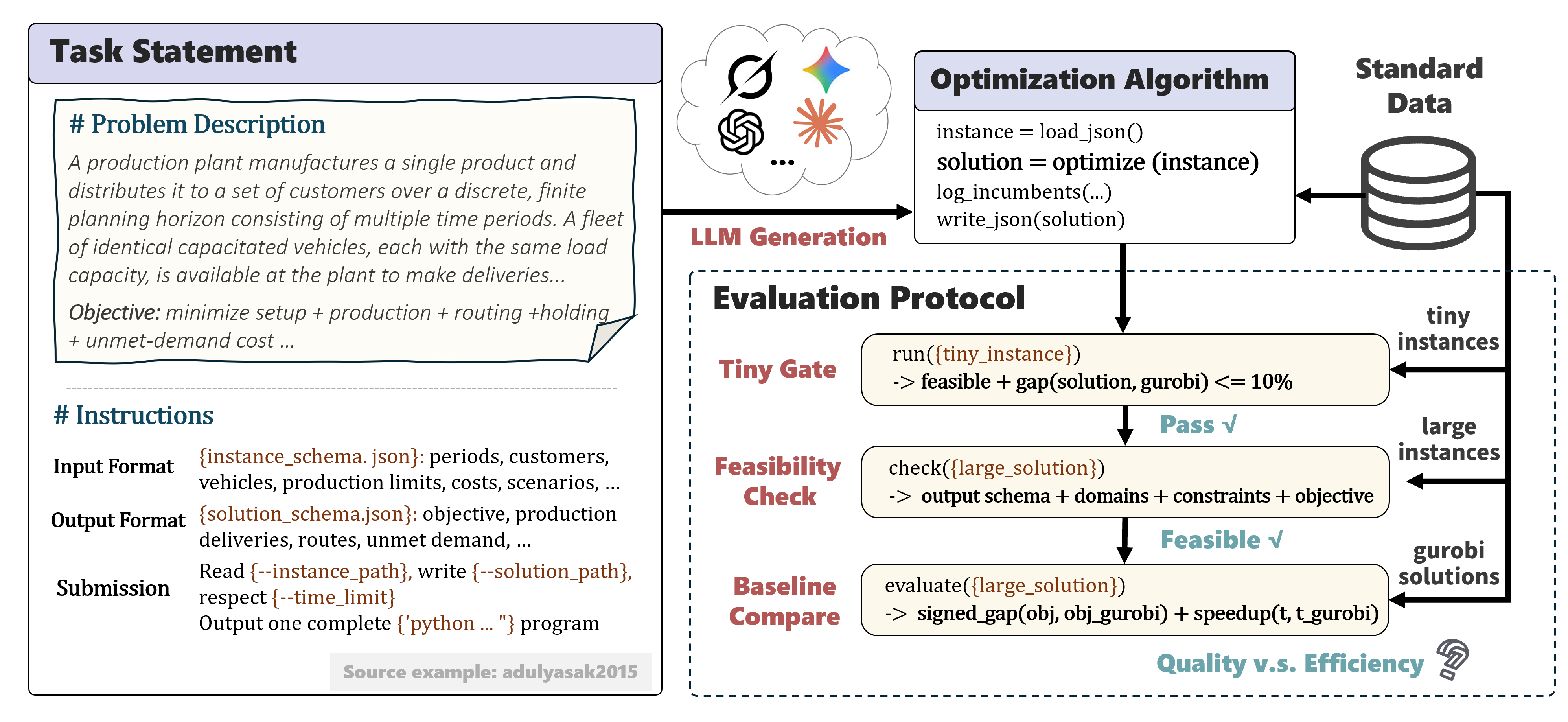}
\caption{Evaluation Protocol of a FrontierOR benchmark task. LLM-generated programs pass a tiny-instance gate before large-instance evaluation of feasibility, solution quality, and efficiency relative to the Gurobi reference.}

\label{fig:eval-pipeline}
\end{figure}

\textbf{Task interface and solution evaluation.}
We evaluate the end-to-end process from problem understanding and formulation recovery to algorithm design and implementation.
For each task, the model receives a natural-language problem
description and input/output schemas and generates a complete
Python program that reads a standardized instance and returns
a solution. The mathematical formulation, Gurobi implementation,
reference solutions, and feasibility checker are withheld from
the model. The task-specific checker validates the returned
solution's required output structure, variable domains, hard
constraints, and objective consistency. Invalid solutions receive no solution-quality or QTE credit; valid solutions are compared with the Gurobi reference in both objective quality and runtime using the metrics defined in Section~\ref{subsec:eval-metrics}.

\textbf{One-shot generation and base models.}
The one-shot generation setting in Figure~\ref{fig:eval-pipeline} evaluates the model's direct capability to infer the underlying optimization structure, choose a solution method, and implement executable code. For each task, the model generates a complete optimization program from scratch. The model performs up to five self-debug iterations for runtime errors only. We introduce a tiny-instance sanity check that rejects programs that time out, return an infeasible solution, or exceed a $10\%$ optimality gap. We use a looser threshold here than the $1\%$ criterion used to measure solution quality on large instances, because it serves only as a preliminary screen to avoid the costly evaluation of unqualified programs at scale.
We evaluate 12 LLMs spanning nine providers: \textit{Claude Fable 5} and \textit{Claude Opus 4.6} (Anthropic); \textit{GPT-5.3-Codex} and \textit{GPT-5.6-Sol} (OpenAI); \textit{Gemini 3.1 Pro Preview} (Google DeepMind); \textit{Kimi K3} (Moonshot AI); \textit{Qwen3.8-Max} and \textit{Qwen3-Coder-Plus} (Alibaba); \textit{DeepSeek-R1} (DeepSeek); \textit{GLM-5.2} (Z.ai); \textit{Grok-4.20} (xAI); and \textit{LLaMA-4-Maverick} (Meta).\footnote{The evaluated model identifiers and initial release dates are claude-fable-5 (released 2026-06-09), claude-opus-4.6 (released 2026-02-04), gpt-5.3-codex (released 2026-02-24), gpt-5.6-sol (released 2026-07-09), gemini-3.1-pro-preview (released 2026-02-19), kimi-k3 (released 2026-07-16), qwen3.8-max (released 2026-08-03), qwen3-coder-plus (released 2025-09-23), deepseek-r1 (released 2025-01-20), glm-5.2 (released 2026-06-16), grok-4.20 (released 2026-03-31), and llama-4-maverick (released 2025-04-05).}

\textbf{Self-evolving frameworks and implementation details.}
The test-time evolution setting evaluates how algorithmic performance
improves through iterative program refinement under a fixed search budget.
We compare three representative frameworks under a unified evaluator while preserving their native feedback mechanisms: (1) \textit{AlphaEvolve} \citep{novikov2025alphaevolve}: The LLM acts as a mutation operator within a MAP-Elites evolutionary algorithm over a program database; we adopt \textit{OpenEvolve} \citep{sharma2025openevolve} as the open-source replication of this paradigm. (2) \textit{EoH} \citep{liu2024evolution}: The LLM jointly evolves code and the natural-language thoughts behind it; its prompt operators explicitly encourage the agent to explore diverse solution approaches rather than locally perturbing existing code. (3) \textit{CORAL} \citep{qu2026CORAL}\footnote{Disclosure: CORAL was developed by a team that includes several authors of this paper. To mitigate potential bias, all frameworks use the same evaluator, candidate budget, initialization, feedback signals, and held-out test protocol.}: In contrast to the single-agent exploration above, the system replaces fixed evolutionary search with autonomous multi-agents that explore, reflect, and collaborate through persistent shared memory.

All frameworks start from the same one-shot program, with a budget of 20 candidate evaluations per task. We use the large instance with the median Gurobi runtime for development and hold out the other four for testing. Each candidate's development solving time is limited to the Gurobi reference runtime, capped at one hour. Candidates are scored by a cascaded fitness function. They first pass the same tiny-instance gate as in one-shot evaluation, then receive a score on the development instance: infeasible or missing solutions score $0$; gaps above $1\%$ earn only quality credit; gaps of at most $1\%$, including solutions better than the reference, also earn a speed bonus based on runtime savings relative to Gurobi. OpenEvolve and EoH receive the signed relative solution gap, runtime relative to Gurobi, and checker violation messages as evolution feedback; CORAL agents can directly inspect the same signals through grader output and evaluation files.

\section{Results}
\label{sec:results}
Our experiments address two research questions: \textbf{(1)} How well do LLMs generate feasible, high-quality, and efficient algorithms in one shot, and what algorithmic patterns and failure modes emerge across models? \textbf{(2)} How much does test-time self-evolution improve these algorithms, and how does performance evolve over iterations?

\subsection{Algorithm Performance with One-Shot Generation}

\label{sec:results-onehot}

We compare twelve base models on FrontierOR's \emph{Full set} ($N=180$) and \emph{Hard subset} ($N=50$) using four metrics in Table~\ref{tab:oneshot-standard-hard}. The Hard subset contains tasks that are potentially more computationally demanding.

\begin{table}[h]
    \caption{One-shot performance on FrontierOR.
    Metric definitions follow Section~\ref{subsec:eval-metrics}. Bold and underlined values indicate the highest and second-highest values in each column, respectively.
    }
    \label{tab:oneshot-standard-hard}
    \centering
    \setlength{\tabcolsep}{6pt}
    \renewcommand{\arraystretch}{1.25}
    \sbox{\FrontierOROneShotTableBox}{%
    \begin{tabular}{lcccc|cccc}
    \toprule
    \textbf{Model} & \multicolumn{4}{c|}{\textbf{FrontierOR Full} ($N=180$)} & \multicolumn{4}{c}{\textbf{FrontierOR Hard} ($N=50$)} \\
    \cmidrule(lr){2-5} \cmidrule(lr){6-9}
     & \textbf{Exec. rate} & \textbf{Feasibility} & \textbf{Sol. quality} & \textbf{QTE} & \textbf{Exec. rate} & \textbf{Feasibility} & \textbf{Sol. quality} & \textbf{QTE} \\
    \midrule
    \rowcolor{FrontierORHighlight}
\textit{Claude Fable 5} & $\mathbf{0.99}$ & $\mathbf{0.69}$ & $\mathbf{0.60}$ & $\mathbf{0.39}$ & $\mathbf{1.00}$ & $\mathbf{0.65}$ & $\mathbf{0.53}$ & $\mathbf{0.51}$ \\
\textit{Claude Opus 4.6} & $\mathbf{0.99}$ & $\underline{0.58}$ & $\underline{0.44}$ & $\underline{0.30}$ & $\mathbf{1.00}$ & $\underline{0.53}$ & $\underline{0.36}$ & $\underline{0.36}$ \\
\textit{GPT-5.3-Codex} & $0.98$ & $0.47$ & $0.38$ & $0.25$ & $\underline{0.98}$ & $0.34$ & $0.24$ & $0.23$ \\
\textit{Gemini 3.1 Pro} & $0.98$ & $0.46$ & $0.42$ & $0.26$ & $0.96$ & $0.34$ & $0.24$ & $0.23$ \\
\textit{Kimi K3} & $0.93$ & $0.39$ & $0.35$ & $0.23$ & $0.90$ & $0.31$ & $0.27$ & $0.24$ \\
\textit{GPT-5.6-Sol} & $0.93$ & $0.34$ & $0.29$ & $0.22$ & $0.92$ & $0.16$ & $0.10$ & $0.09$ \\
\textit{Qwen3.8-Max} & $\underline{0.99}$ & $0.44$ & $0.32$ & $0.21$ & $\mathbf{1.00}$ & $0.36$ & $0.23$ & $0.22$ \\
\textit{DeepSeek-R1} & $0.94$ & $0.28$ & $0.22$ & $0.16$ & $\underline{0.98}$ & $0.15$ & $0.10$ & $0.10$ \\
\textit{GLM-5.2} & $0.92$ & $0.30$ & $0.22$ & $0.14$ & $0.90$ & $0.22$ & $0.11$ & $0.10$ \\
\textit{Grok-4.20} & $0.94$ & $0.21$ & $0.18$ & $0.12$ & $0.92$ & $0.08$ & $0.05$ & $0.05$ \\
\textit{Qwen3-Coder-Plus} & $0.87$ & $0.17$ & $0.13$ & $0.08$ & $0.84$ & $0.08$ & $0.05$ & $0.05$ \\
\textit{LLaMA-4-Maverick} & $0.81$ & $0.09$ & $0.08$ & $0.04$ & $0.82$ & $0.05$ & $0.03$ & $0.03$ \\
    \bottomrule
    \end{tabular}%
    }
    \global\FrontierOROneShotNaturalWidth=\wd\FrontierOROneShotTableBox
    \resizebox{\textwidth}{!}{\usebox{\FrontierOROneShotTableBox}}
    \vspace{-8pt}
    \end{table}

\raggedbottom
\afterpage{\flushbottom}
\textbf{Performance across models.}
Table~\ref{tab:oneshot-standard-hard} highlights four patterns.
(1) \textit{Correct and scalable solving, rather than executability, remains the bottleneck.}
Full-set execution rates reach $0.98$--$0.99$ for the leading models and remain above $0.80$ for all models. Yet feasibility ranges from only $0.09$ to $0.69$, and solution quality from $0.08$ to $0.60$. Errors in constraint specification and output formatting cause tiny-gate failures, while larger instances further expose difficulties in constructing and refining feasible solutions and finding high-quality solutions within the time budget.
(2) \textit{Solving efficiency remains less competitive than advanced solvers.}
QTE shows that only $4\%$--$39\%$ of Full-set large instances are solved at least as fast as the Gurobi reference, with feasible solutions whose objective gap relative to the reference is at most $1\%$. The efficiency gap reflects limitations in formulation design, search strategy and termination criteria, as well as implementation overhead in LLM-generated programs.
(3) \textit{Comparison between the Full set and Hard subset.}
All twelve models lose feasibility and solution quality on Hard, by $3.8$--$17.8$ and $5.6$--$19.4$ percentage points, respectively. Nevertheless, Fable 5, Opus 4.6, Kimi K3, and Qwen3.8-Max attain higher QTE on Hard than on Full. This reflects the computational difficulty of Hard tasks: unlike its strong runtime performance on easier tasks, Gurobi approaches runtime saturation on nearly all Hard instances. This leaves greater scope for LLM-generated programs to match its solution quality in less time.
(4) \textit{Model performance and cost-effectiveness.}
Premium frontier models Fable 5 and Opus 4.6 lead, with Full-set QTE of $0.39$ and $0.30$, whereas the earlier, lower-cost Qwen3-Coder-Plus and LLaMA-4-Maverick achieve only $0.08$ and $0.04$. However, cost-effective models such as DeepSeek-R1 and Qwen3.8-Max also deliver competitive performance, reaching QTE of $0.16$ and $0.21$, respectively.
% We provide continuous measures of solution quality and quality-time efficiency in Appendix~\ref{app:continuous-metrics}.
Appendix~\ref{app:source-paper-algorithms} reevaluates LLM-generated programs using specialized algorithms from the source papers as the alternative reference.
Appendix~\ref{app:formulation-given-ablation} reports an ablation with and without the mathematical formulation as input to assess whether performance gaps largely stem from formulation recovery or downstream algorithm design.
Appendix~\ref{app:case-studies-failures} presents case studies illustrating the successes and failures of LLM-generated solver programs in terms of correctness, solution quality, and computational efficiency.

\begin{figure}[H]
\centering
\includegraphics[width=0.8\linewidth]{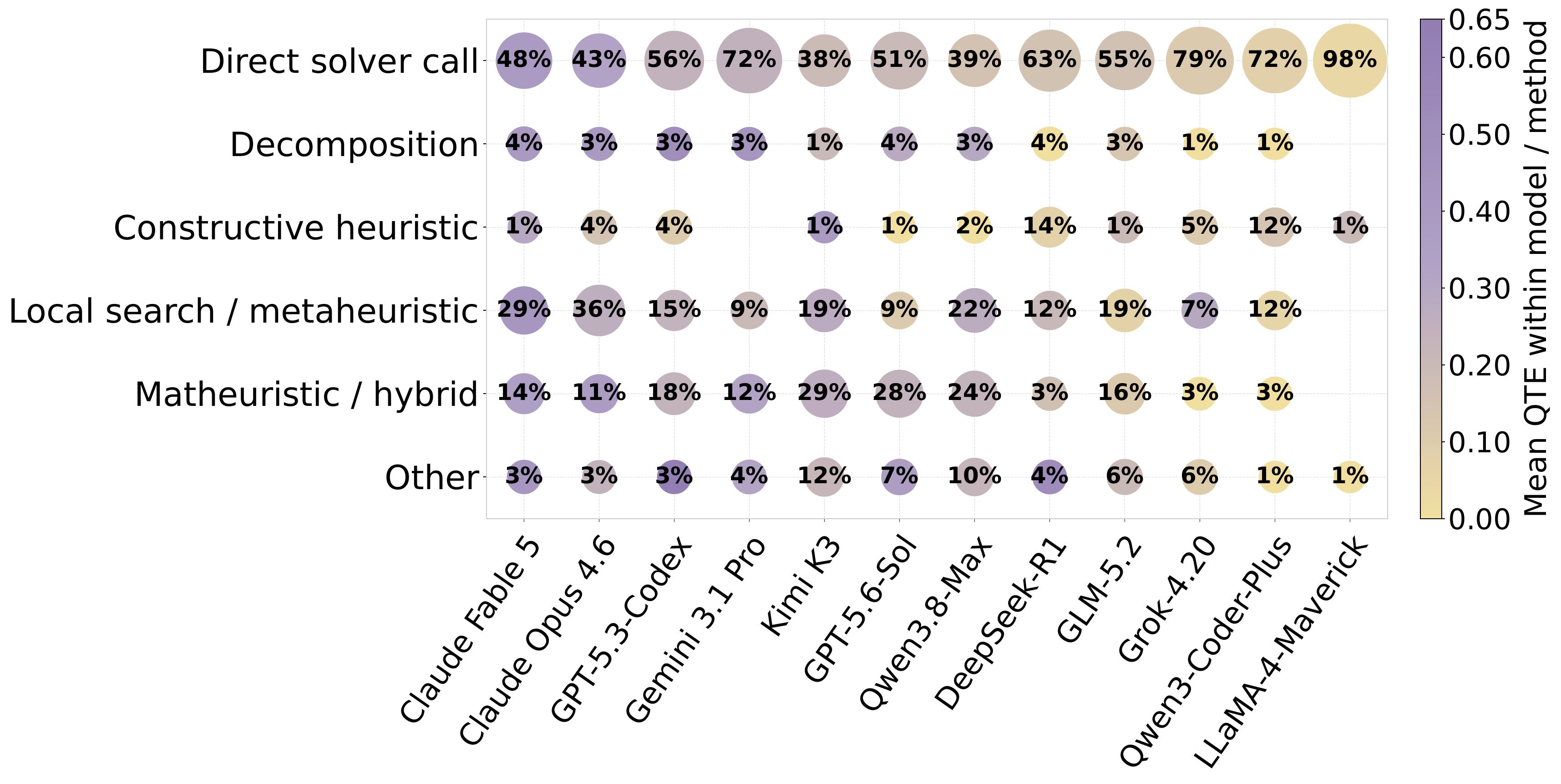}
\caption{Solution methods across LLMs. Bubble size and labels indicate category proportions per model; color shows mean QTE within each model–category group.}
\label{fig:alg-family-bubble}
\label{fig:per_model_break_down}
\end{figure}

\textbf{Solution methods across models.}
Figure~\ref{fig:alg-family-bubble} reveals a contrast between solver-centric and more diverse algorithm designs. Earlier models such as Qwen3-Coder-Plus and LLaMA-4-Maverick rely heavily on direct solver calls, whereas several newer models, including Claude Fable 5, Claude Opus 4.6, Kimi K3, and Qwen3.8-Max, more frequently employ local-search, metaheuristic, or hybrid methods. This trend suggests that stronger models select solution methods more flexibly to suit problem structure, helping balance solution quality and efficiency and thereby improve QTE.

\begin{figure}[H]
\centering
\includegraphics[width=0.8\linewidth]{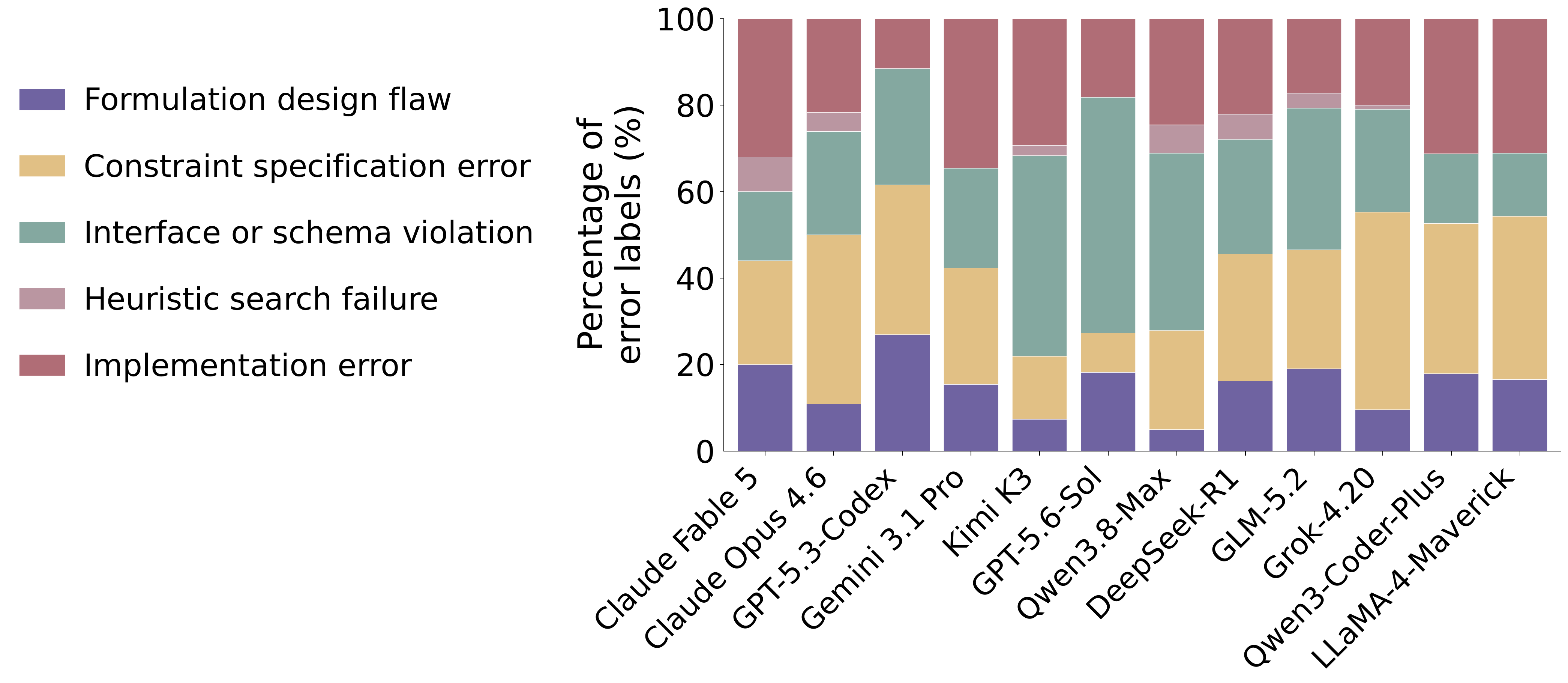}
\caption{Failure-mode composition across LLMs. Each segment represents the proportion of one error category in the model’s total label count across the five categories.}
\label{fig:failure-mode-stacked}
\end{figure}

\textbf{Failure modes in generated solver programs.}
Figure~\ref{fig:failure-mode-stacked} shows that constraint-specification errors and interface/schema violations are the most common error types, followed by implementation errors, while heuristic-search failures account for a small share. Typical issues include omitted capacity or flow-conservation constraints, incorrect output fields or indexing, and programming defects such as undefined variables or invalid array accesses. Constraint-specification errors dominate the error composition of Grok-4.20, Qwen3-Coder-Plus, and LLaMA-4-Maverick, accounting for $34.8\%$--$45.7\%$ of their error annotations. Stronger models exhibit different error compositions: constraint-specification errors remain the largest category for Claude Opus 4.6 ($39.1\%$), whereas implementation errors are the largest category for Gemini 3.1 Pro and Claude Fable 5 ($34.6\%$ and $32.0\%$, respectively).
The classification procedures and examples for algorithmic patterns and failure modes are detailed in Appendix~\ref{app:classification-taxonomy}.

\subsection{Performance Improvement with Test-time Evolution}
\label{sec:self-evolve-results}

Direct one-shot generation, even with the strongest models, leaves substantial room for improvement on the Hard subset (Table~\ref{tab:oneshot-standard-hard}). We compare three
self-evolving frameworks with GPT-5.6-Sol as their shared
LLM backend. For each task, all frameworks start from the
same one-shot program, generated by the model achieving
the highest average one-shot QTE across the Hard subset.

\begingroup
\setlength{\intextsep}{5pt}
\begin{table}[H]
\color{black}
\centering
\definecolor{evolutiongainbg}{HTML}{EEEAF9}
\setlength{\tabcolsep}{4pt}
\caption{Dev and Test performance on the Hard subset. Parentheses report percentage-point changes from the one-shot program. Here, $n$ denotes the number of task instances.}
\label{tab:self-evolve-test}
\resizebox{\linewidth}{!}{%
\begin{tabular}{lrrrrrr}
\toprule
 & \multicolumn{3}{c}{Dev ($n=50$)} & \multicolumn{3}{c}{Test ($n=200$)} \\
\cmidrule(lr){2-4}\cmidrule(lr){5-7}
Method & Feas. & SQ & QTE & Feas. & SQ & QTE \\
\midrule
Starter & $0.66$ & $0.52$ & $0.50$ & $0.65$ & $0.54$ & $0.51$ \\
OpenEvolve & $0.96$ \colorbox{evolutiongainbg}{\strut\scriptsize $(+0.30)$} & $\underline{0.96}$ \colorbox{evolutiongainbg}{\strut\scriptsize $(+0.44)$} & $\underline{0.96}$ \colorbox{evolutiongainbg}{\strut\scriptsize $(+0.46)$} & $0.89$ \colorbox{evolutiongainbg}{\strut\scriptsize $(+0.24)$} & $0.76$ \colorbox{evolutiongainbg}{\strut\scriptsize $(+0.23)$} & $0.76$ \colorbox{evolutiongainbg}{\strut\scriptsize $(+0.25)$} \\
EoH & $\underline{0.98}$ \colorbox{evolutiongainbg}{\strut\scriptsize $(+0.32)$} & $0.94$ \colorbox{evolutiongainbg}{\strut\scriptsize $(+0.42)$} & $0.94$ \colorbox{evolutiongainbg}{\strut\scriptsize $(+0.44)$} & $\underline{0.93}$ \colorbox{evolutiongainbg}{\strut\scriptsize $(+0.28)$} & $\underline{0.77}$ \colorbox{evolutiongainbg}{\strut\scriptsize $(+0.23)$} & $\underline{0.76}$ \colorbox{evolutiongainbg}{\strut\scriptsize $(+0.25)$} \\
CORAL & $\textbf{1.00}$ \colorbox{evolutiongainbg}{\strut\scriptsize $(+0.34)$} & $\textbf{0.98}$ \colorbox{evolutiongainbg}{\strut\scriptsize $(+0.46)$} & $\textbf{0.98}$ \colorbox{evolutiongainbg}{\strut\scriptsize $(+0.48)$} & $\textbf{0.96}$ \colorbox{evolutiongainbg}{\strut\scriptsize $(+0.31)$} & $\textbf{0.80}$ \colorbox{evolutiongainbg}{\strut\scriptsize $(+0.27)$} & $\textbf{0.78}$ \colorbox{evolutiongainbg}{\strut\scriptsize $(+0.27)$} \\
\bottomrule
\end{tabular}}
\end{table}

\endgroup
\paragraph{Self-evolution improvements across development and test sets.}
All three self-evolving frameworks improve over the one-shot baseline across every metric (Table~\ref{tab:self-evolve-test}). CORAL attains the highest point estimate on all six Dev and Test metrics, consistent with its autonomous multi-agent design, which can flexibly explore and combine candidate methods rather than rely on predefined evolutionary workflows. OpenEvolve performs slightly better on the development set in SQ and QTE ($0.96$ vs. $0.94$), whereas EoH performs slightly better on the test set in feasibility and SQ ($0.93$ vs. $0.89$ and $0.77$ vs. $0.76$), with QTE tied at $0.76$. EoH's slightly stronger Test performance contrasts with OpenEvolve's higher Dev scores. Appendix~\ref{app:statistical-uncertainty} reports task-level bootstrap confidence intervals for self-evolution performance and paired comparisons.

\begingroup
\setlength{\intextsep}{3pt}
\begin{figure}[H]
\centering
\includegraphics[width=\textwidth]{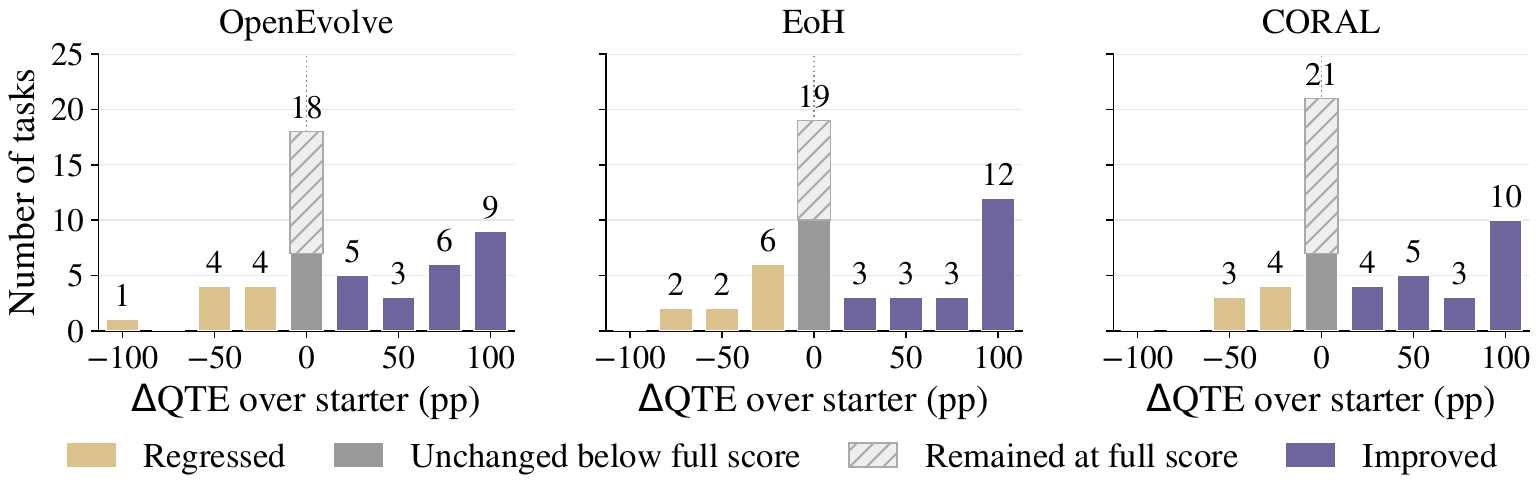}
\caption{Per-task QTE improvements over the starter program on four held-out Test instances. Horizontal axes show final score changes in percentage points. Hatched zero-change segments denote tasks remaining at full score; solid gray denotes unchanged scores below one.}
\label{fig:self-evolve-task-improvement}
\end{figure}
\endgroup

\textbf{Task-level improvements and regressions.}
Figure~\ref{fig:self-evolve-task-improvement} shows that QTE improves on 23, 21, and 22 tasks for OpenEvolve, EoH, and CORAL, respectively, but declines on 9, 10, and 7. CORAL achieves the largest average QTE gain with the fewest task-level regressions, while EoH records slightly more regressions than OpenEvolve. Regressions on held-out instances indicate that development improvements do not always generalize. 
Appendix~\ref{app:case-studies} examines improvement, regression, and unchanged cases.

\section{Conclusion}
\label{sec:conclusion}
We present \textbf{FrontierOR}, a benchmark for evaluating whether LLMs can move beyond correct formulation to efficient algorithm design for realistic large-scale OR problems. The benchmark requires models to infer problem structure from natural-language specifications and generate algorithms that compete with human-verified Gurobi baselines in feasibility, solution quality and efficiency.

Our evaluation shows that current LLMs still face substantial challenges in designing correct, scalable, and efficient optimization algorithms. Despite high execution rates, generated programs struggle to maintain feasibility and solution quality as instance size grows. Even the strongest one-shot model achieves only $39\%$ quality--time efficiency (QTE) against the Gurobi reference on the Full set, owing to LLMs' weaknesses in formulation choice, algorithm design, and efficient implementation.  
Analysis examines both solution strategies and failure modes: stronger models tailor solution methods to problem structure, improving quality--time efficiency. Constraint-specification errors and interface/schema violations are the most common failure types, followed by implementation errors.
Test-time evolution frameworks improve substantially by refining candidate programs with evaluation feedback, although gains on development instances do not always carry over to held-out instances. Further research should explore evolution strategies with stronger cross-instance generalization.

Admittedly, FrontierOR's task coverage is limited by the source OR literature, and its empirical findings are limited to the recently released models evaluated in this study and the experimental settings. For each task, FrontierOR includes a small set of fixed instances, which may not capture the full range of instance distributions or changing operational environments. Metrics such as solution quality and quality--time efficiency (QTE) evaluate the final output solution under a fixed time budget, without comparing algorithms' broader anytime performance. Nevertheless, FrontierOR provides a foundation and evaluation testbed for developing LLMs and self-evolving systems for scalable and efficient optimization algorithm design.

\section*{Acknowledgments}
This research is supported by the National Research Foundation (NRF), Prime Minister's Office, Singapore, under its Campus for Research Excellence and Technological Enterprise (CREATE) programme. The Mens, Manus, and Machina (M3S) is an interdisciplinary research group (IRG) of the Singapore--MIT Alliance for Research and Technology (SMART) centre.

% \newpage
\bibliographystyle{plainnat}
\bibliography{main}

\appendix

\raggedbottom

\section{Detailed Benchmark Construction Protocol}
\label{app:construction-details}

\subsection{Source Selection and Reproduction Pipeline}
\label{app:source-selection}

For each selected paper, the optimization problem, test data, and reference solution methods are extracted and reproduced through a three-stage pipeline:
\begin{enumerate}
    \item \textbf{Formulation extraction.} The original mathematical formulation of the optimization problem---sets, parameters, variables, objective, and constraints---is identified from the paper and transcribed in \LaTeX{} as the single authoritative reference for construction and verification. When the paper presents multiple alternative formulations, we select the formulation that is most central to the computational study. When the paper explicitly gives a solver-ready reformulation, we also record it as an implementation reference.
    \item \textbf{Instance specification extraction.} The configurations of all test-instance sets used in the paper's computational experiments are extracted into a machine-readable JSON specification. When instances rely on external and accessible benchmarks, their configurations are retrieved and cross-checked to reproduce the scale and structure used in the source paper.
    \item \textbf{Solver implementation.} Based on the solver-ready formulation, we implement a Gurobi program that returns solutions expressed in the original decision variables. If the solver uses auxiliary reformulation variables internally, the output is projected back to the variables of the original formulation. When the paper omits a solver-ready formulation, we derive one from the mathematical model and document any necessary assumptions.
\end{enumerate}

\subsection{Natural-Language Problem Description}
\label{app:problem-description-protocol}

The problem description is the primary input presented to the LLMs under evaluation. It conveys all information necessary to design a correct solution method, without revealing mathematical formulas or algorithmic hints. Each description is synthesized from three sources: (i) the original paper, which provides the application context and operational semantics; (ii) the original mathematical formulation, which canonically defines the modeling rules; and (iii) the instance and solution schemas, which anchor the input and output formats.

The description follows the following style constraints:
\begin{itemize}
    \item The text is continuous natural-language prose. No \LaTeX{} formulas, numbered lists, section headers, or structural markup are included in the LLM-facing specification.
    \item Mathematical relationships are verbalized rather than written symbolically; for example, we write \emph{each machine serves at most one product per period} rather than $\sum_i x_{ijt} \leq 1$.
    \item Optimization-modeling jargon such as \emph{decision variable}, \emph{constraint}, or \emph{objective function} is replaced by operational language, e.g., \emph{the company must decide...} or \emph{the goal is to minimize...}.
    \item Common-sense mathematical properties, such as non-negativity of physical quantities or integrality of yes/no decisions, are omitted unless they constitute essential modeling requirements that cannot be inferred from the business context.
\end{itemize}

\subsection{Test Instances and Ground Truth}
\label{app:instances-ground-truth}

Each task includes two categories of test instances. \textbf{Tiny instances} are minimal-scale instances, typically solvable by Gurobi in seconds, that serve as a lightweight functional gate: They verify that an LLM-generated program executes, parses input and output correctly, and can produce at least one feasible solution before consuming computational budget on large instances. \textbf{Real-world large instances} are the primary scoring targets. Their scale is determined by expert reasoning grounded in the operational context and by the experimental configurations of the source paper. Where possible, we diversify structural characteristics at comparable nominal scales---for example, spatial layouts, graph topologies, or capacity-to-load ratios---because such changes can alter Gurobi runtime by orders of magnitude even when the MIP size is similar.

The Gurobi baseline is obtained by running the human-verified solver implementation of the mathematical formulation within the benchmark time budget. It yields a certified optimal objective when Gurobi proves optimality and the best feasible objective otherwise. The output includes the full solution vector, objective value, solver status, and solve time. A standalone Python feasibility checker verifies every hard constraint, variable domain, and variable bound against a candidate solution and reports any violation and its magnitude. To prevent LLMs from reporting an objective inconsistent with their submitted solution, every checker also re-computes the objective from the reported decision variables and uses that re-computed value.

\section{Quality-Assurance Details}
\label{app:qa-details}

\paragraph{Automated cross-verification.}
The feasibility checker is run against every Gurobi solution on every instance. Any reported violation triggers a diagnosis-and-repair loop. Observed issues fall into two broad categories: (i) \emph{checker-side errors}, such as overly tight numerical tolerances, misimplemented constraints, or mismatched parsing logic between the checker and the solver; and (ii) \emph{solver-side errors}, such as omitted constraints, incorrect objective terms, or ill-scaled formulations in the Gurobi code. Solver-side fixes require re-solving the affected instances. After iterative repair, all retained Gurobi solutions pass the feasibility check.

\paragraph{Expert review and re-verification.}
We engaged 15 PhD researchers in operations research and related quantitative fields who, over a three-week period, independently audited the complete component suite of each paper---mathematical formulation, problem description, Gurobi code, and feasibility checker implementation---against the original publication. Each task was assigned to a primary reviewer whose research area matches its problem class, with the 180 tasks distributed uniformly across the 15 reviewers (approximately 12 tasks per reviewer). The review followed a structured checklist that covered:
\begin{itemize}
    \item \textbf{Mathematical formulation fidelity:} whether the extracted model is correct and complete, with all sets, parameters, variables, objective terms, and constraints faithfully extracted without omissions or redundancies.
    \item \textbf{Problem-description quality:} whether every element of the mathematical model has a corresponding natural-language description that is complete, unambiguous, and does not admit multiple reasonable interpretations.
    \item \textbf{Gurobi-code alignment:} whether the solver implementation is strictly consistent with the mathematical formulation or with an explicit reformulation used for solver tractability, and whether unavoidable defaults such as penalty coefficients or big-$M$ values are reasonable.
    \item \textbf{Feasibility-checker scope:} whether the checker verifies exactly the hard constraints, variable domains, and variable bounds of the mathematical model, while excluding soft constraints, penalties, and heuristic rules.
\end{itemize}
The main errors flagged by experts concerned symbol and indexing errors in the mathematical formulation, incomplete or ambiguous natural-language descriptions, missing constraints or unauthorized simplifications in the solver code, and vacuous or incomplete feasibility checks. On average, each task went through two to three rounds of repair and re-verification before all four checklist items passed, after which we obtained a clean final benchmark version.

\subsection{Sensitivity to the Problem-Description Source}
\label{app:construction-bias}

This paired analysis examines whether using Claude Code (Opus 4.6) to generate task components introduces a performance advantage for Claude-family models through the problem-description wording. In one-shot generation, the description supplies the model-facing task narrative, while the formulation and evaluators remain hidden. We therefore replace only the original Claude-generated descriptions with Codex-regenerated versions, keeping all other task components and evaluation settings unchanged.

We select 20 tasks on which Claude Opus 4.6 achieves high large-instance QTE with the original descriptions. We evaluate Claude Opus 4.6 to test whether its performance declines when the original descriptions are replaced with Codex-generated descriptions.
Table~\ref{tab:description-source-bias} shows replacing Claude-generated descriptions with Codex-generated descriptions increases Opus 4.6's execution rate, feasibility, and SQ by $5$, $1$, and $2$ percentage points, respectively, while QTE decreases by $5$ points. The absence of a consistent decline across metrics suggests that its performance in that experiment was not uniformly dependent on Claude-generated wording.

\begin{table}[!htbp]
\centering
\begingroup\small
\caption{Description-source comparison for Claude Opus 4.6.}
\label{tab:description-source-bias}
\setlength{\tabcolsep}{5pt}
\begin{tabular}{llcccc}
\toprule
Model & Description source & Exec. & Feas. & SQ & QTE \\
\midrule
\textit{Claude Opus 4.6} & Claude (original) & $0.63$ & $0.73$ & $0.59$ & $0.53$ \\
 & Codex & $0.68$ & $0.74$ & $0.61$ & $0.48$ \\
\bottomrule
\end{tabular}
\endgroup
\end{table}

\subsection{Paper-Memory Audit and Contamination Analysis}
\label{app:contamination}

Although task descriptions are rewritten and expert-reviewed to omit paper titles, authors, citations, and unnecessary paper-specific notation or algorithm names, models may still recognize published work from its problem setting. We therefore conduct a conservative, title-only audit: each model receives a source-paper title, but not its abstract or full text, and is asked to describe the problem and algorithmic design. An independent Codex judge with access to the paper PDF assesses whether the response contains accurate, paper-specific algorithmic details unlikely to be inferred from the title alone. A positive result is evidence of potential paper-specific familiarity, not proof that the model used memorized information when solving the benchmark; a negative result likewise does not establish that contamination is absent.

Positive evidence was limited for most models: eight of the eleven models other than Claude Fable 5 were flagged on no more than 17 papers each; Claude Opus 4.6, GPT-5.6-Sol, and Gemini 3.1 Pro were flagged on 25, 29, and 30, respectively. Claude Fable 5 had the highest count, 103 papers (57.2\%). However, after excluding the tasks flagged for Claude Fable 5, its feasibility, solution quality, and QTE change only from 0.69, 0.60, and 0.39 to 0.67, 0.57, and 0.42, respectively. The QTE increase is modest, suggesting that these tasks have only a limited effect on its overall quality--time performance. More broadly, even within model-specific flagged-task subsets, performance was not uniformly successful: feasibility, solution quality, and QTE reached at most 0.80, 0.62, and 0.47, respectively. These results suggest that recalling parts of a source algorithm is not enough: reliable performance still requires understanding the problem structure and translating it into an appropriate, correctly implemented algorithm.

To estimate how much these flags affect aggregate results, we remove every task flagged for at least one model and recompute all models on the same remaining audited-task subset (74 of 180 tasks; Table~\ref{tab:paper-memory-audit}). Excluding flagged tasks changes feasibility, solution quality, and QTE by up to 0.08 (8.3 percentage points), 0.07 (7.5 points), and 0.06 (5.9 points), respectively. Although some middle-ranked models exchange positions, the leaders in solution quality and QTE remain unchanged. This sensitivity analysis suggests that the main comparative conclusions are not driven by the tasks flagged in this audit.

\begin{table}[htbp]
\centering
\begingroup
\color{black}
\small
\setlength{\tabcolsep}{4pt}
\caption{Paper-memory evidence and performance before and after excluding potentially memorized tasks. Clean means excluding any task with positive memorization evidence for at least one model.}
\label{tab:paper-memory-audit}
\begin{tabular}{lcccc}
\toprule
Model & \shortstack{Papers with\\memory evidence} & \shortstack{Feas.\\(full $\to$ clean)} & \shortstack{SQ\\(full $\to$ clean)} & \shortstack{QTE\\(full $\to$ clean)} \\
\midrule
Claude Fable 5 & 103 & 0.69 $\to$ 0.66 & 0.60 $\to$ 0.57 & 0.39 $\to$ 0.42 \\
Claude Opus 4.6 & 25 & 0.58 $\to$ 0.55 & 0.44 $\to$ 0.41 & 0.30 $\to$ 0.30 \\
GPT-5.3-Codex & 6 & 0.47 $\to$ 0.39 & 0.38 $\to$ 0.31 & 0.25 $\to$ 0.22 \\
Gemini 3.1 Pro & 30 & 0.46 $\to$ 0.42 & 0.42 $\to$ 0.39 & 0.26 $\to$ 0.26 \\
Kimi K3 & 17 & 0.39 $\to$ 0.41 & 0.35 $\to$ 0.37 & 0.23 $\to$ 0.27 \\
GPT-5.6-Sol & 29 & 0.34 $\to$ 0.35 & 0.29 $\to$ 0.32 & 0.22 $\to$ 0.22 \\
Qwen3.8-Max & 8 & 0.44 $\to$ 0.39 & 0.32 $\to$ 0.25 & 0.21 $\to$ 0.15 \\
DeepSeek-R1 & 6 & 0.28 $\to$ 0.20 & 0.22 $\to$ 0.15 & 0.16 $\to$ 0.11 \\
GLM-5.2 & 7 & 0.30 $\to$ 0.28 & 0.22 $\to$ 0.18 & 0.14 $\to$ 0.11 \\
Grok-4.20 & 9 & 0.21 $\to$ 0.15 & 0.18 $\to$ 0.12 & 0.12 $\to$ 0.09 \\
Qwen3-Coder-Plus & 0 & 0.17 $\to$ 0.13 & 0.13 $\to$ 0.09 & 0.08 $\to$ 0.07 \\
LLaMA-4-Maverick & 3 & 0.09 $\to$ 0.06 & 0.08 $\to$ 0.04 & 0.04 $\to$ 0.03 \\
\bottomrule
\end{tabular}
\endgroup
\end{table}

\section{Additional Dataset Statistics}
\label{app:dataset-statistics}
This appendix complements the overview in Section~\ref{subsec:statistics} with detailed statistics on the benchmark's publication venues, temporal distribution (year of publication), and algorithmic diversity (the families of solution methods proposed by the source papers), shown in Table~\ref{tab:venues-algos}.

\begin{table}[h]
\caption{Publication venues and algorithmic approaches of the 180 source papers. Algorithmic approaches refer to the methods proposed in the source papers.}
\label{tab:venues-algos}
\centering
\small
\begin{tabular}{l r@{\quad} l r}
\toprule
\textbf{Venue} & \textbf{\%} &
\textbf{Algorithmic Approach} & \textbf{\%} \\
\midrule
INFORMS Journal on Computing              & 31 & Branch-and-Price / Column Generation              & 30 \\
European Journal of Operational Research   & 17 & Branch-and-Cut                                    & 18 \\
Transportation Science                     & 13 & Metaheuristics / Matheuristics                    & 13 \\
Operations Research                        & 13 & Benders Decomposition                             & 12 \\
Computers \& Operations Research           &  9 & Tight Formulation / Other Exact                   & 10 \\
Management Science                         &  3 & Branch-and-Bound (Mixed-Integer / Semidefinite)    & 9 \\
Mathematical Programming / Computation     &  1 & Dynamic Programming / Decision Diagrams            &  4 \\
Other                                      & 12 & Other Decomposition (Lagrangian, ADMM, etc.)       &  3 \\
\bottomrule
\end{tabular}
\end{table}

\paragraph{Temporal distribution.}
The collection spans 1992--2025: 33 papers (18\%) predate 2010, 19 papers (11\%) were published in 2010--2014, 59 papers (33\%) in 2015--2019, and 69 papers (38\%) in 2020 or later.

\paragraph{Algorithmic diversity.}
Branch-and-price / column generation (30\%) and branch-and-cut (18\%) are the two most prevalent paradigms. Decomposition methods---Benders decomposition (12\%), Lagrangian relaxation (2\%), and ADMM / stochastic decomposition ($<$1\%)---together account for $15\%$ of the collection. Metaheuristics and matheuristics (13\%) and dynamic-programming or decision-diagram methods (4\%) add further diversity in algorithm-design style.

\section{Supplementary Analyses of Algorithmic Performance}
\label{app:supplementary-quality-analysis}

This section examines the robustness of the main experimental results in Section~\ref{sec:results} and how LLM-generated algorithm performance varies with reference solution methods and input context.

\subsection{Sensitivity to the Solution-Gap Threshold}
\label{app:gap-threshold-sensitivity}

\begin{figure}[!h]
    \centering
    \includegraphics[width=\textwidth]{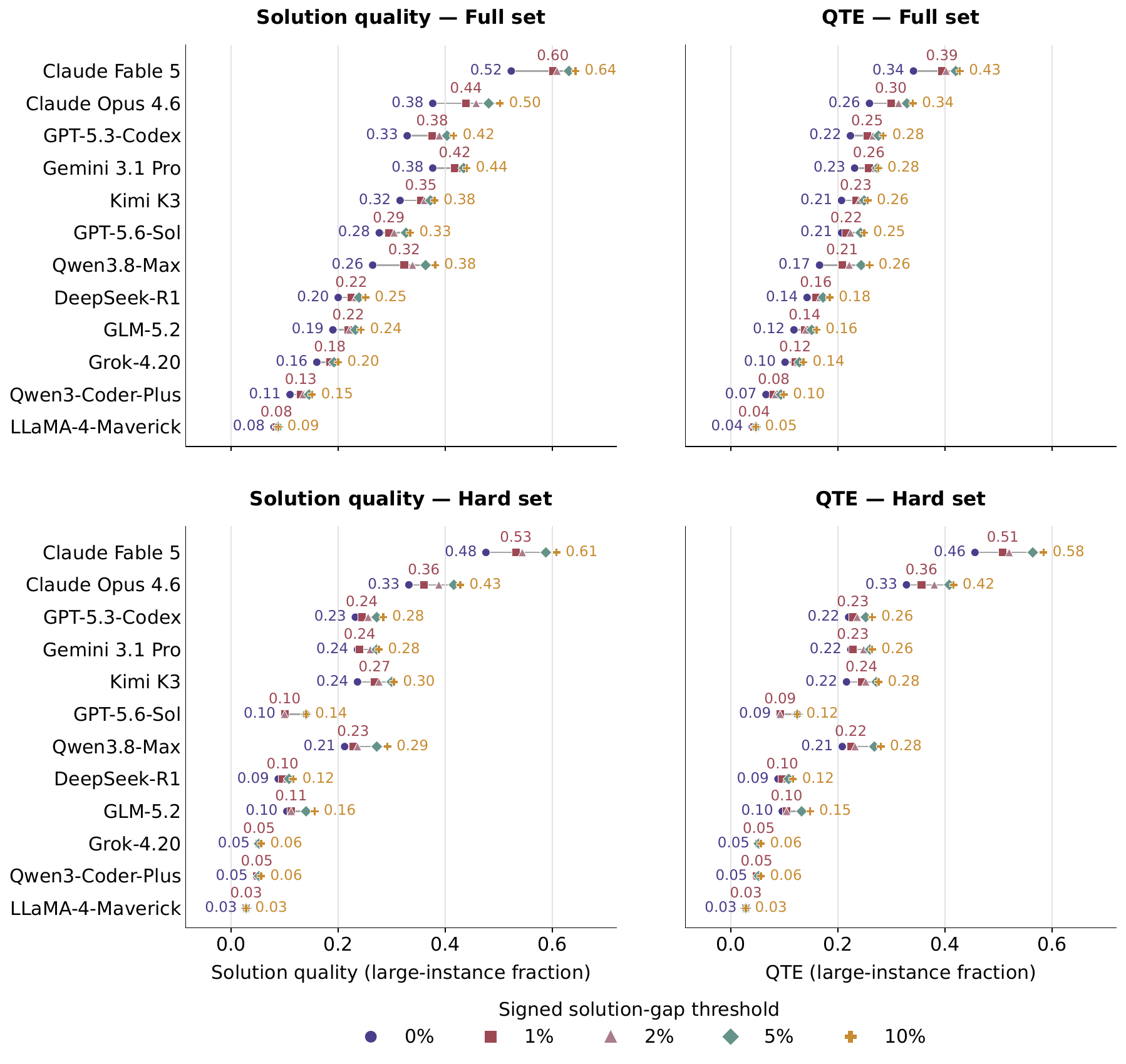}
    \caption{Sensitivity of solution quality and QTE to the signed relative solution-gap threshold on the Full set and Hard subset. Within each model, markers from left to right correspond to increasingly permissive thresholds from $0\%$ to $10\%$.}
    \label{fig:gap-threshold-sensitivity}
\end{figure}

The main evaluation uses a $1\%$ signed relative solution-gap threshold. To assess whether the conclusions depend on this choice, we recompute solution quality and QTE on the Full set and Hard subset at thresholds of $0\%$, $1\%$, $2\%$, $5\%$, and $10\%$. Figure~\ref{fig:gap-threshold-sensitivity} connects the five values for each model; the endpoint labels report the results at $0\%$ and $10\%$, with the main $1\%$ setting also annotated.

Both metrics are non-decreasing as the threshold is relaxed. Claude Fable 5 and Claude Opus 4.6 consistently rank first and second, respectively, in both solution quality and QTE on both sets. From $0\%$ to $10\%$, solution quality increases by $0.8$--$12.6$ percentage points on Full and $0.0$--$13.2$ points on Hard; QTE increases by $0.7$--$9.3$ and $0.0$--$12.8$ points, respectively. Averaged across models, the gains are smaller for QTE than for solution quality: $5.1$ versus $6.9$ points on Full, and $4.6$ versus $5.0$ points on Hard. The smaller QTE gains reflect the additional runtime requirement, which some newly admitted high-quality solutions do not satisfy. Some middle-ranked models exchange positions: for example, GPT-5.3-Codex trails Gemini 3.1 Pro in Full-set QTE at $0\%$ ($0.223$ versus $0.231$), but leads at $10\%$ ($0.284$ versus $0.276$). Thus, the leading models and the overall performance gap remain robust, although the complete ranking is not invariant. Nevertheless, even at $10\%$, the highest QTE is only $0.428$ on Full and $0.584$ on Hard.

\subsection{Comparison with Reproduced Source-Paper Algorithms}
\label{app:source-paper-algorithms}

The monolithic Gurobi reference is a general formulate-and-solve baseline, whereas the source papers often present specialized algorithms designed by OR experts for their problem structures. Comparing LLM-generated algorithms with these methods therefore provides a more direct assessment of the gap between LLM and human-expert algorithm design. We implemented all 180 reproductions ourselves in Python, using Gurobi where applicable and adapting each implementation to FrontierOR's instance and output requirements. Some source methods were originally implemented in other languages or relied on solver-specific or proprietary features, so substitutions or adaptations were sometimes necessary and may limit reproduction fidelity.
Against Gurobi, the reproduced algorithms achieve SQ/QTE of $0.79/0.63$ on the Full set and $0.70/0.67$ on the Hard subset. This means that they attain a solution within $1\%$ of or better than the Gurobi reference on $79\%$ of instances in the Full set and $70\%$ in the Hard subset, respectively, and do so in no greater time on $63\%$ and $67\%$ of instances, respectively.

Table~\ref{tab:expert-baseline-provisional} reports SQ and QTE recomputed using the reproduced source-paper algorithms as references, with all other evaluation settings unchanged.
\begin{table}[h]
\centering
\begingroup\color{black}
\caption{One-shot performance using source-paper specialized algorithms as references. Parentheses show relative percentage changes from the scores computed against the Gurobi reference in Table~\ref{tab:oneshot-standard-hard}. Best and second-best scores are bold and underlined.}
\label{tab:expert-baseline-provisional}
\setlength{\tabcolsep}{6pt}
\renewcommand{\arraystretch}{1.25}
% Match Table 3's actual font size, not merely its font commands.
\resizebox{\textwidth}{!}{%
\begin{tabular*}{\FrontierOROneShotNaturalWidth}{@{\extracolsep{\fill}}lcc|cc}
\toprule
\textbf{Model} & \multicolumn{2}{c|}{\textbf{FrontierOR Full} ($N=180$)} & \multicolumn{2}{c}{\textbf{FrontierOR Hard} ($N=50$)} \\
\cmidrule(lr){2-3}\cmidrule(lr){4-5}
 & \textbf{Sol. quality} & \textbf{QTE} & \textbf{Sol. quality} & \textbf{QTE} \\
\midrule
\rowcolor{FrontierORHighlight}
\textit{Claude Fable 5} & $\mathbf{0.58}\;(-2.8\%)$ & $\mathbf{0.22}\;(-44.8\%)$ & $\mathbf{0.51}\;(-3.8\%)$ & $\mathbf{0.20}\;(-61.4\%)$ \\
\textit{Claude Opus 4.6} & $0.39\;(-10.4\%)$ & $0.17\;(-41.6\%)$ & $\underline{0.26}\;(-26.7\%)$ & $\underline{0.15}\;(-58.4\%)$ \\
\textit{GPT-5.3-Codex} & $0.35\;(-6.5\%)$ & $\underline{0.18}\;(-30.1\%)$ & $0.18\;(-24.6\%)$ & $0.07\;(-70.2\%)$ \\
\textit{Gemini 3.1 Pro} & $\underline{0.40}\;(-3.7\%)$ & $0.16\;(-39.7\%)$ & $0.22\;(-6.7\%)$ & $0.04\;(-80.7\%)$ \\
\textit{Kimi K3} & $0.35\;(-1.9\%)$ & $0.17\;(-26.2\%)$ & $0.25\;(-7.5\%)$ & $0.10\;(-59.0\%)$ \\
\textit{GPT-5.6-Sol} & $0.29\;(-1.9\%)$ & $0.15\;(-28.9\%)$ & $0.10\;(-4.0\%)$ & $0.06\;(-30.4\%)$ \\
\textit{Qwen3.8-Max} & $0.30\;(-6.2\%)$ & $0.16\;(-24.6\%)$ & $0.17\;(-26.3\%)$ & $0.12\;(-44.6\%)$ \\
\textit{DeepSeek-R1} & $0.20\;(-10.9\%)$ & $0.11\;(-30.8\%)$ & $0.04\;(-58.3\%)$ & $0.03\;(-66.7\%)$ \\
\textit{GLM-5.2} & $0.21\;(-4.6\%)$ & $0.11\;(-20.2\%)$ & $0.10\;(-14.3\%)$ & $0.04\;(-61.5\%)$ \\
\textit{Grok-4.20} & $0.18\;(-1.2\%)$ & $0.11\;(-12.0\%)$ & $0.04\;(-30.8\%)$ & $0.02\;(-53.8\%)$ \\
\textit{Qwen3-Coder-Plus} & $0.13\;(-3.4\%)$ & $0.07\;(-13.9\%)$ & $0.04\;(-25.0\%)$ & $0.02\;(-50.0\%)$ \\
\textit{LLaMA-4-Maverick} & $0.08\;(-2.6\%)$ & $0.05\;(+5.1\%)$ & $0.03\;(+0.0\%)$ & $0.02\;(-28.6\%)$ \\
\bottomrule
\end{tabular*}}
\vspace{-8pt}
\endgroup
\end{table}
Replacing the Gurobi references with the reproduced algorithms lowers mean Full-set QTE across the twelve models from $0.200$ to $0.137$, an absolute decrease of $0.063$ ($31.4\%$). QTE declines for eleven models, by $12.0\%$--$44.8\%$; only LLaMA-4-Maverick increases slightly, from $0.043$ to $0.046$. This consistent decline shows that the reproduced source-paper methods provide a more demanding baseline for evaluating LLM-generated algorithm design than monolithic Gurobi.

\subsection{Comparison with Formulation as Additional Input}
\label{app:formulation-given-ablation}

The standard natural-language setting evaluates problem understanding, formulation recovery, algorithm design, and implementation jointly. To assess whether the observed performance gaps largely stem from formulation recovery or downstream algorithm design and implementation, we compare inputs with and without the mathematical formulation.

We select 50 tasks whose mathematical formulations contain no explicit algorithmic hints. We evaluate five high-performing one-shot models---Claude Fable 5, Claude Opus 4.6, GPT-5.3-Codex, Gemini 3.1 Pro, and Kimi K3---to examine whether supplying the mathematical formulation alleviates the performance bottlenecks that persist even among strong models. The augmented prompt appends the mathematical formulation to the standard problem description. All other evaluation settings remain unchanged. Table~\ref{tab:formulation-given-ablation} compares natural-language-only input (using the corresponding results from Table~\ref{tab:oneshot-standard-hard}) with natural language plus the mathematical formulation.

\begin{table}[htbp]
\centering
\caption{Comparison with and without the mathematical formulation as additional input. NL denotes the standard natural-language input; NL+F additionally includes the formulation. Mean denotes the average across the five models.}
\label{tab:formulation-given-ablation}
\setlength{\tabcolsep}{6pt}
\renewcommand{\arraystretch}{1.25}

\resizebox{\textwidth}{!}{%
{\color{black}
\begin{tabular*}{\FrontierOROneShotNaturalWidth}{@{\extracolsep{\fill}}lcc|cc|cc}
\toprule
 & \multicolumn{2}{c}{\textbf{Feas.}} & \multicolumn{2}{c}{\textbf{SQ}} & \multicolumn{2}{c}{\textbf{QTE}} \\
\cmidrule(lr){2-3}\cmidrule(lr){4-5}\cmidrule(lr){6-7}
\textbf{Model} & NL & NL+F & NL & NL+F & NL & NL+F \\
\midrule
\textit{Claude Fable 5} & $0.62$ & $0.68$ & $0.50$ & $0.60$ & $0.34$ & $0.44$ \\
\textit{Claude Opus 4.6} & $0.54$ & $0.49$ & $0.38$ & $0.35$ & $0.26$ & $0.22$ \\
\textit{GPT-5.3-Codex} & $0.42$ & $0.31$ & $0.37$ & $0.29$ & $0.28$ & $0.20$ \\
\textit{Gemini 3.1 Pro} & $0.44$ & $0.45$ & $0.39$ & $0.44$ & $0.29$ & $0.30$ \\
\textit{Kimi K3} & $0.41$ & $0.40$ & $0.35$ & $0.35$ & $0.26$ & $0.23$ \\
\midrule
Mean & $0.49$ & $0.46$ & $0.40$ & $0.41$ & $0.29$ & $0.28$ \\
\bottomrule
\end{tabular*}}}
\end{table}

Providing the formulation changes mean feasibility from $0.487$ to $0.464$, SQ from $0.399$ to $0.406$, and QTE from $0.286$ to $0.275$. Mean SQ improves slightly, but feasibility and QTE decline on average. Changes vary across models: Fable 5 and Gemini 3.1 Pro gain $9.2$ and $0.8$ percentage points in QTE, whereas Claude Opus 4.6, GPT-5.3-Codex, and Kimi K3 decline by $4.4$, $7.6$, and $3.2$ points, respectively. Even with the mathematical formulation, LLMs may choose or implement algorithms in ways that yield infeasible solutions or incur additional computational costs, so providing the formulation may not consistently improve solution quality or runtime efficiency.

\subsection{Statistical Uncertainty}
\label{app:statistical-uncertainty}
\begingroup\color{black}

We quantify uncertainty in held-out test-time evolution performance due to task composition using 50,000 task-level bootstrap resamples and report percentile 95\% confidence intervals (CIs) in Table~\ref{tab:evolution-bootstrap}. Each resample draws the original number of tasks with replacement, keeping each task's evaluation results together. For each framework, we also compute its QTE minus the starter's QTE on the same resampled tasks. The resulting 95\% CIs for these paired differences are positive for all three frameworks (OpenEvolve: $[0.105, 0.385]$; EoH: $[0.105, 0.395]$; CORAL: $[0.140, 0.395]$), supporting the observed gains from test-time evolution. The frameworks achieve broadly comparable QTE, with some uncertainty in their relative ordering. The intervals also indicate that the magnitude of improvement can vary with task composition.

\begin{table}[!htbp]
\centering\color{black}\small
\caption{Held-out performance on Hard subset, with task-level bootstrap 95\% confidence intervals.}
\label{tab:evolution-bootstrap}
\setlength{\tabcolsep}{5pt}
\begin{tabular}{lccc}
\toprule
Method & Feas. & SQ & QTE \\
\midrule
Starter & $0.650\;[0.520, 0.775]$ & $0.535\;[0.410, 0.660]$ & $0.510\;[0.385, 0.635]$ \\
OpenEvolve & $0.890\;[0.805, 0.960]$ & $0.760\;[0.670, 0.840]$ & $0.755\;[0.665, 0.835]$ \\
EoH & $0.930\;[0.875, 0.975]$ & $0.765\;[0.685, 0.840]$ & $0.760\;[0.675, 0.840]$ \\
CORAL & $0.955\;[0.905, 0.990]$ & $0.800\;[0.715, 0.875]$ & $0.775\;[0.685, 0.860]$ \\
\bottomrule
\end{tabular}
\end{table}
\endgroup

\subsection{Continuous Measures}
\label{app:continuous-metrics}
The metrics \emph{solution quality} and \emph{QTE} reported in Table~\ref{tab:oneshot-standard-hard} are thresholded binary pass rates: a (paper, instance) case counts as a pass only if the generated program returns a feasible solution with an objective gap of at most $1\%$ relative to the Gurobi reference (and, for \emph{QTE}, at least as fast as the reference). These 0/1 indicators are easy to aggregate but discard magnitude information. We complement them with continuous comparisons of final solution quality and total solving time. For each task--instance pair, the subject program $X$ is compared with a reference program $Y$, either Gurobi or another LLM-generated program.

To compute the continuous solution quality $\Delta q$, we compare the final objective values $o_X(T),o_Y(T)$ returned under the common time budget $T=1$ hour. Let $\delta\in\{+1,-1\}$ the optimization direction ($+1$ for minimization, $-1$ for maximization). We use a $\max$-normalized signed difference, bounded in $[-2,2]$ in general and in $[-1,1]$ when the objectives have the same sign; when both objectives are zero, we set $\Delta q=0$:
\begin{equation}
\label{eq:delta-q}
\Delta q \;=\; \delta\cdot\frac{o_Y(T)-o_X(T)}{\max\!\left(|o_X(T)|,\,|o_Y(T)|\right)}\;\in\;[-2,2].
\end{equation}
For continuous solving-time advantage $\Delta t$, let $t_X$ and $t_Y$ denote the complete programs' evaluator-recorded wall-clock runtimes. The logarithm centers the runtime ratio at zero, and the arctangent transformation bounds the score within $(-1,1)$, with positive values indicating shorter solving time for $X$.
\begin{equation}
\label{eq:delta-t}
\Delta t \;=\; \phi\!\left(\frac{\widetilde t_Y}{\widetilde t_X}\right),
\qquad \widetilde t_Z=\max(t_Z,10^{-3}\,\mathrm{s}),\quad Z\in\{X,Y\},
\qquad \phi(r)=\tfrac{2}{\pi}\arctan(\log r).
\end{equation}
The score lies in $(-1,1)$; positive values indicate shorter total solving time for $X$. We use recorded runtimes without budget capping, retaining minor execution overheads. This metric measures the advantage in total solving time and should be interpreted alongside final solution quality to characterize quality--time trade-offs.

\subsubsection{Comparison with the Gurobi Reference}
\label{app:per-task-ratios}

Table~\ref{tab:per-task-ratios} reports all twelve models using the current evaluation results, Gurobi references, and Full/Hard partition. Quality means include checker-feasible final outputs with finite objectives. Time means additionally require the final output to meet the same signed $1\%$ quality threshold as the main evaluation, including the existing absolute tolerances for certified-zero references. They compare total program runtimes against Gurobi's recorded runtime. Negative $\overline{\Delta q}$ indicates worse average final solution quality than Gurobi on feasible instances. Negative $\overline{\Delta t}$ indicates a tendency toward longer total solving times than Gurobi on quality-qualified instances.

\begin{table}[!h]
\caption{Continuous final solution quality and total solving-time advantage relative to Gurobi. Numbers in parentheses indicate the number of instances used to compute each mean. Quality is averaged over instances with feasible final solutions. Time additionally requires final quality to meet the main evaluation threshold. Higher is better in both columns.}
\label{tab:per-task-ratios}
\centering
\setlength{\tabcolsep}{6pt}
\small
\begin{tabular}{l rr | rr}
\toprule
\textbf{Model / Method} & \multicolumn{2}{c|}{\textbf{FrontierOR Full} ($n=180$)} & \multicolumn{2}{c}{\textbf{FrontierOR Hard} ($n=50$)} \\
\cmidrule(lr){2-3} \cmidrule(lr){4-5}
 & $\overline{\Delta q}$ & $\overline{\Delta t}$ & $\overline{\Delta q}$ & $\overline{\Delta t}$ \\
\midrule
\textit{Claude Fable 5} & $0.038$ (601) & $-0.021$ (541) & $0.110$ (161) & $0.114$ (133) \\
\textit{Claude Opus 4.6} & $-0.010$ (514) & $0.046$ (395) & $0.052$ (126) & $0.218$ (90) \\
\textit{GPT-5.3-Codex} & $0.000$ (422) & $0.135$ (338) & $0.121$ (82) & $0.198$ (61) \\
\textit{Gemini 3.1 Pro} & $0.000$ (416) & $0.041$ (376) & $-0.031$ (84) & $0.140$ (60) \\
\textit{Kimi K3} & $0.048$ (355) & $0.157$ (319) & $0.167$ (78) & $0.225$ (67) \\
\textit{GPT-5.6-Sol} & $0.026$ (305) & $0.369$ (265) & $0.111$ (38) & $0.628$ (25) \\
\textit{Qwen3.8-Max} & $-0.004$ (392) & $0.197$ (291) & $0.088$ (90) & $0.489$ (57) \\
\textit{DeepSeek-R1} & $-0.004$ (251) & $0.278$ (202) & $0.140$ (37) & $0.763$ (24) \\
\textit{GLM-5.2} & $-0.054$ (252) & $0.159$ (197) & $0.023$ (46) & $0.504$ (28) \\
\textit{Grok-4.20} & $-0.016$ (190) & $0.224$ (166) & $-0.056$ (20) & $0.537$ (13) \\
\textit{Qwen3-Coder-Plus} & $-0.023$ (150) & $0.239$ (117) & $0.000$ (19) & $0.768$ (12) \\
\textit{LLaMA-4-Maverick} & $0.016$ (84) & $0.134$ (76) & $-0.302$ (12) & $0.771$ (7) \\
\bottomrule
\end{tabular}
\end{table}

The results reveal three patterns.
(1) \textit{Average quality on feasible instances is close to Gurobi, with larger differences across models on Hard.} Full-set $\overline{\Delta q}$ ranges from $-0.054$ to $0.048$ and is positive for 6 of the twelve models. Thus, when programs return feasible solutions, their average quality is often close to Gurobi's. This complements the main results: low overall solution-quality pass rates do not necessarily imply poor average quality among feasible outputs; consistently producing feasible solutions remains an important bottleneck. On Hard, the range widens to $-0.302$--$0.167$, revealing larger differences on difficult tasks.
(2) \textit{Most models have a total-solving-time advantage when final quality qualifies.} Full-set $\overline{\Delta t}$ is positive for eleven models, and all twelve have positive means on Hard. Programs that meet the reference-quality requirement therefore show potential to reduce solving time, especially on difficult tasks where Gurobi exhibits solver saturation. This is consistent with low overall QTE: QTE also counts failures due to infeasibility or insufficient quality, whereas these time means consider quality-qualified instances. GPT-5.6-Sol achieves a mean time advantage of $0.369$, whereas Fable 5, despite leading aggregate QTE and a mean quality advantage of $0.038$, has a time advantage of $-0.021$, illustrating a possible time cost of better quality.
(3) \textit{Strong conditional quality is not restricted to the highest-ranked models.} Kimi K3 achieves the highest mean quality advantage on both Full ($0.048$) and Hard ($0.167$). On Hard, DeepSeek-R1 and Qwen3.8-Max also achieve positive mean quality advantages of $0.140$ and $0.088$. These means use different feasible or quality-qualified instance subsets for each model. They reveal potential in successful cases rather than an overall ranking; the next section examines differences through pairwise comparisons on common instances.

\subsubsection{Pairwise Comparison across Models}
\label{app:pair-wise-omparison}

We compare all $66$ model pairs on the shared large task--instance pairs $S_{AB}$ where both programs return checker-feasible solutions with finite objectives, thereby excluding infeasible or invalid outputs from the comparison. We average final solution quality and total solving time over the same instances for each pair; positive values favor column model $A$ over row model $B$ on the corresponding dimension.

\begin{figure}[!ht]
\centering
\begin{subfigure}[t]{0.50\textwidth}
    \centering
    \includegraphics[width=\linewidth]{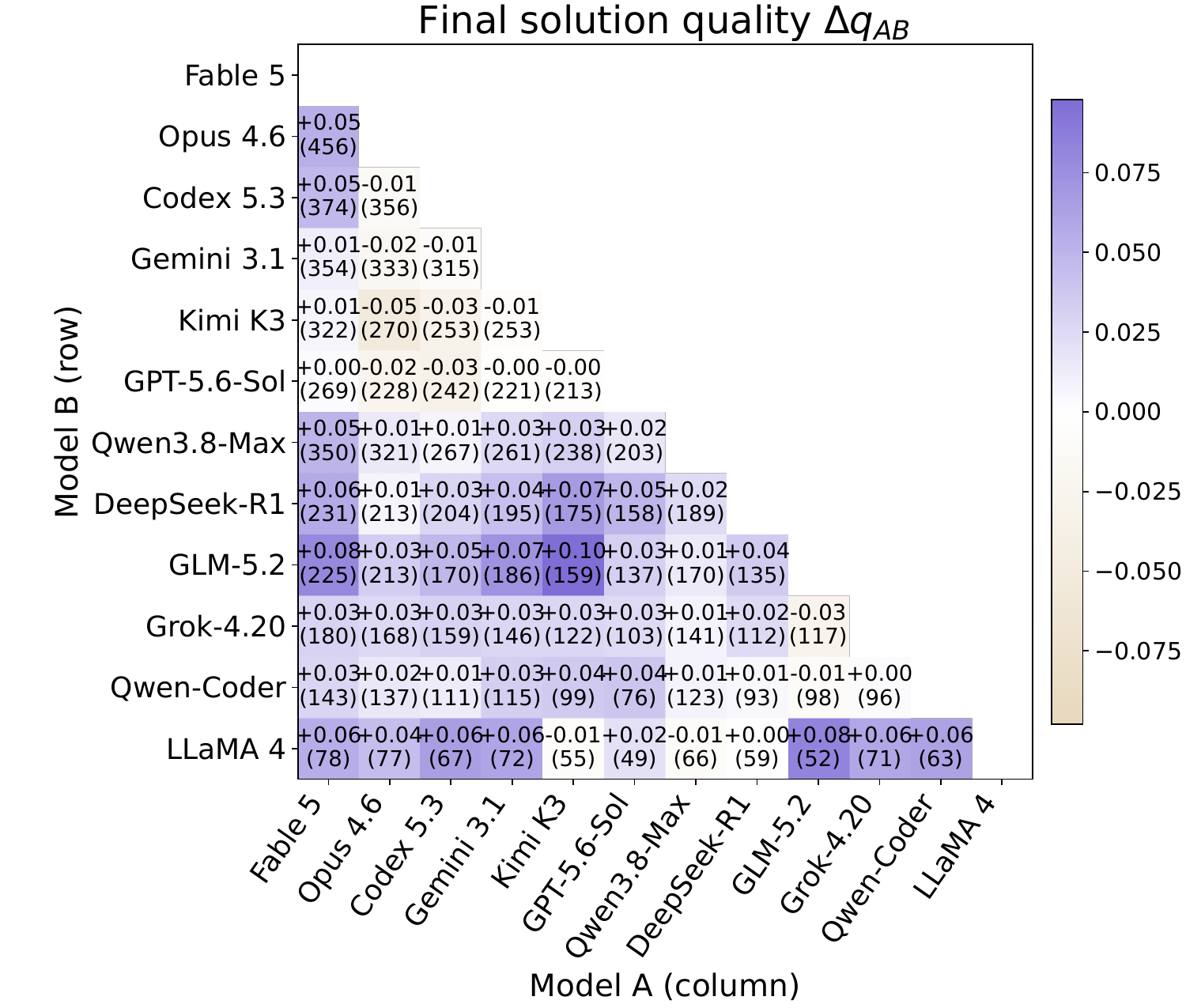}
    \caption{Final solution quality.}
    \label{fig:pairwise-quality-heatmap}
\end{subfigure}\hfill
\begin{subfigure}[t]{0.50\textwidth}
    \centering
    \includegraphics[width=\linewidth]{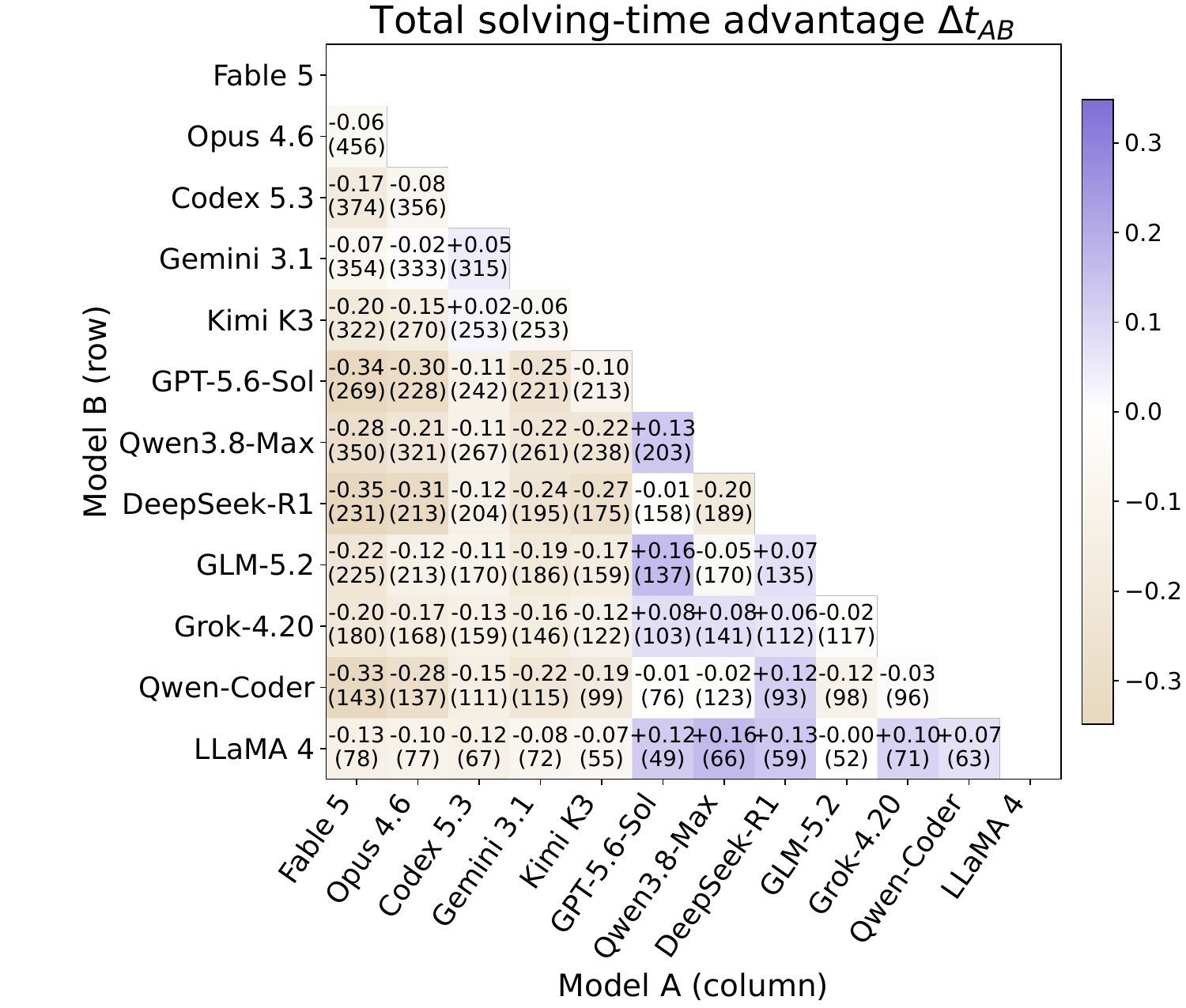}
    \caption{Total solving-time advantage.}
    \label{fig:pairwise-time-heatmap}
\end{subfigure}
\caption{Pairwise comparison across twelve models on shared feasible Full-set large instances. Each lower-triangular cell compares column model $A$ against row model $B$; positive (purple) values favor $A$ on the corresponding dimension. Parentheses give the shared instance count.}
\label{fig:pairwise-heatmap}
\end{figure}

Figure~\ref{fig:pairwise-heatmap} shows three patterns.
(1) \textit{Solution-quality gaps narrow on shared feasible instances.} Of the $66$ Full-set model pairs, $53$ have $|\overline{\Delta q}_{AB}|<0.05$, and the largest magnitude is $0.098$. Although LLaMA-4-Maverick ranks lowest overall in Table~\ref{tab:oneshot-standard-hard}, its conditional solution quality is close to DeepSeek-R1 and slightly higher than Kimi K3 and Qwen3.8-Max (by about $0.6$ and $0.9$ percentage points), while Fable 5, Opus 4.6, GPT-5.3-Codex, and Gemini 3.1 Pro lead it by about $4$--$6$ points. LLaMA's programs predominantly use monolithic Gurobi formulations ($151$ of $152$ programs with algorithm-family labels), which can yield competitive solutions on successfully modeled cases but does not ensure broad feasibility or task coverage.
(2) \textit{Runtime comparisons reveal different quality--time trade-offs.} On shared feasible instances, Fable 5 achieves higher average solution quality than Opus 4.6 ($\overline{\Delta q}_{AB}=0.053$) but takes longer overall ($\overline{\Delta t}_{AB}=-0.060$). GPT-5.3-Codex is slightly behind Gemini 3.1 Pro in quality ($-0.009$) but faster in total runtime ($+0.046$). LLaMA is also faster than the four leading models on their shared feasible instances ($\overline{\Delta t}_{AB}=-0.080$ to $-0.125$ when each leader is column model $A$), despite its lower conditional solution quality. These examples show that quality and total runtime do not yield a uniform model ordering.
(3) \textit{Aggregate differences partly reflect solution coverage.} Differences in execution and feasibility rates widen the overall SQ and QTE gaps between algorithms generated by different models. On instances where both programs return feasible solutions, solution-quality gaps narrow considerably, while solving times exhibit model-dependent trade-offs, with some lower-ranked models finishing faster than leading models.

% \newpage
\section{Task Examples and Case studies}
\label{app:dataset-example}

% \clearpage

\subsection{Classification of Solution Methods and Failure Modes}
\label{app:classification-taxonomy}

To calibrate the solution-method and failure-mode annotation protocols, an OR expert independently reviewed 50 cases for each taxonomy, sampled across its categories. Errors identified in this review were used to refine the classification criteria and add corrective instructions to the agent-audit prompts. Agents then applied the revised protocols to complete the full annotation. For solution-method annotation, an agent inspects each generated program and assigns exactly one label based on its primary solution strategy, rather than its initialization. The six categories are direct solver call, decomposition, constructive heuristic, local search/metaheuristic, matheuristic/hybrid, and other. For failure-mode annotation, agents inspect the program together with task specifications, checker diagnostics, and saved evaluation results. Each confirmed erroneous program receives one primary root-cause label and any supported secondary labels. The five displayed categories are formulation-design flaws, constraint-specification errors, interface/schema violations, heuristic-search failures, and implementation errors (programming defects distinct from modeling or interface errors).

\textbf{Solution-method examples.} The following cards show concise code excerpts highlighting the program structures used to determine each category.

\noindent
\begin{minipage}[t]{0.49\textwidth}
\vspace{0pt}
\begin{tcolorbox}[taskpane={Direct solver call}, fonttitle=\bfseries\fontsize{9.6}{11.5}\selectfont]
\textit{Example:} \emph{adulyasak2015} / LLaMA-4-Maverick
\begin{lstlisting}[style=taskcode]
model = gp.Model("Production_Distribution")
y = model.addVars(T, vtype=GRB.BINARY)
x = model.addVars(n+1, n+1, m, T,
                  vtype=GRB.INTEGER)
P = model.addVars(T, n_scenarios)
I = model.addVars(n+1, T, n_scenarios)
model.addConstrs(...)  # full formulation
model.optimize()
\end{lstlisting}
\begin{tcolorbox}[codecue]
All scenario-expanded variables and constraints enter one model followed by one solve.
\end{tcolorbox}
\end{tcolorbox}
\end{minipage}\hfill
\begin{minipage}[t]{0.49\textwidth}
\vspace{0pt}
\begin{tcolorbox}[taskpane={Decomposition}, fonttitle=\bfseries\fontsize{9.6}{11.5}\selectfont]
\textit{Example:} \emph{brandao2016} / GPT-5.6-Sol
\begin{lstlisting}[style=taskcode]
master = gp.Model("bin_packing_master")
for _ in range(max_iterations):
    master.optimize()
    duals = [c.Pi for c in constraints]
    pricing = gp.Model("pricing")
    # Integer knapsack; item profits are duals
    pricing.optimize()
    # Collect new patterns with dual value > 1
    if not new_keys: break
    for key in new_keys:
        add_master_variable(key)
\end{lstlisting}
\begin{tcolorbox}[codecue]
The evaluated solver alternates between a restricted master LP and dual-driven knapsack pricing, then solves a restricted integer master.
\end{tcolorbox}
\end{tcolorbox}
\end{minipage}

\medskip
\noindent
\begin{minipage}[t]{0.49\textwidth}
\vspace{0pt}
\begin{tcolorbox}[taskpane={Constructive heuristic}, fonttitle=\bfseries\fontsize{9.6}{11.5}\selectfont]
\textit{Example:} \emph{bode2012} / Qwen3-Coder-Plus
\begin{lstlisting}[style=taskcode]
for vehicle in range(num_vehicles):
    for edge in remaining_required_edges:
        if load + edge["demand"] <= capacity:
            route.append(edge["edge_id"])
            load += edge["demand"]

while unserviced:
    next_edge = unserviced[0]
    visit_closer_endpoint(next_edge)
    unserviced.remove(next_edge)
\end{lstlisting}
\begin{tcolorbox}[codecue]
Greedy packing and routing commit decisions once, with no improvement phase.
\end{tcolorbox}
\end{tcolorbox}
\end{minipage}\hfill
\begin{minipage}[t]{0.49\textwidth}
\vspace{0pt}
\begin{tcolorbox}[taskpane={Local search / metaheuristic}, fonttitle=\bfseries\fontsize{9.6}{11.5}\selectfont]
\textit{Example:} \emph{alba2013} / DeepSeek-R1 (local search)
\begin{lstlisting}[style=taskcode]
tour = nearest_neighbor_tsp(depot, vertices, dist)
improvement = True
while improvement:
    improvement = False
    for i in range(1, n-1):
        for j in range(i+1, n):
            if new_edge_cost < old_edge_cost:
                tour[i:j+1] = reversed(tour[i:j+1])
                improvement = True
\end{lstlisting}
\begin{tcolorbox}[codecue]
Nearest neighbor only initializes an iterative 2-opt neighborhood search.
\end{tcolorbox}
\end{tcolorbox}
\end{minipage}

\medskip
\noindent
\begin{minipage}[t]{0.49\textwidth}
\vspace{0pt}
\begin{tcolorbox}[taskpane={Local search / metaheuristic}, fonttitle=\bfseries\fontsize{9.6}{11.5}\selectfont]
\textit{Example:} \emph{alba2013} / Claude Opus~4.6 (metaheuristic)
\begin{lstlisting}[style=taskcode]
temperature = best_cost * 0.05 if best_cost > 0 else 100
while time.time() - start_time < time_limit - 1.0:
    new_pt = perturb_tour(current_pickup_tour,
                          pickup_dist)
    new_cost = tour_cost(new_pt, pickup_dist) + ...
    delta = new_cost - current_cost
    accept = delta < 0
    if not accept:
        accept = random.random() < math.exp(-delta / temperature)
    if accept and check_lifo_feasibility(new_po, new_do):
        current_pickup_tour, current_cost = new_pt, new_cost
    temperature *= 0.9995
\end{lstlisting}
\begin{tcolorbox}[codecue]
The primary loop is simulated annealing, with random neighborhood moves, probabilistic acceptance, and temperature decay.
\end{tcolorbox}
\end{tcolorbox}
\end{minipage}\hfill
\begin{minipage}[t]{0.49\textwidth}
\vspace{0pt}
\begin{tcolorbox}[taskpane={Matheuristic / hybrid}, fonttitle=\bfseries\fontsize{9.6}{11.5}\selectfont]
\textit{Example:} \emph{amaldi2013} / Gemini~3.1~Pro
\begin{lstlisting}[style=taskcode]
while time.time() - start_time < time_limit:
    if incumbent:
        pool = get_iteration_w_pool(chosen_subsets, ...)
    else:
        pool = get_initial_w_pool(points, ...)
    global_subsets_list.extend(
        make_clusters(pool))
    k, chosen = solve_set_cover(global_subsets_list,
                                remaining_time,
                                last_sol_indices)
    last_sol_indices = chosen
\end{lstlisting}
\begin{tcolorbox}[codecue]
Heuristic candidate generation repeatedly feeds a solver-based restricted set-cover subproblem.
\end{tcolorbox}
\end{tcolorbox}
\end{minipage}

\textbf{Failure-mode examples.} The following cards pair concise code excerpts with evaluator diagnostics and source-supported root causes for the five categories.

\noindent
\begin{minipage}[t]{0.49\textwidth}
\vspace{0pt}
\begin{tcolorbox}[taskpane={Formulation-design flaw}, fonttitle=\bfseries\fontsize{9.6}{11.5}\selectfont]
\textit{Example:} \emph{arslan2021} / DeepSeek-R1
\begin{lstlisting}[style=taskcode]
r = model.addVars(n, lb=0, name="r")
s = model.addVars(n, lb=0, name="s")
w = model.addVar(lb=0, name="w")
obj += gp.quicksum(r[i] for i in range(n))
obj += uncertainty_budget * w
model.setObjective(obj)  # s is not penalized
for i, item in enumerate(items):
    max_deg = item['max_degradation']
    model.addConstr(r[i] + s[i] + w >=
                    max_deg * (y[i] - z[i]))
\end{lstlisting}
\begin{tcolorbox}[failuremessage]
Tiny objective consistency fails: reported $-9274.33$, versus $-8385.74$ from the checker's exact fixed-$x$ LP relaxation.
\end{tcolorbox}
\begin{tcolorbox}[failurecause]
Unbounded, unpenalized slack $s_i$ can absorb all degradation exposure with $r_i=w=0$, removing the intended uncertainty penalty and changing the robust model.
\end{tcolorbox}
\end{tcolorbox}
\end{minipage}\hfill
\begin{minipage}[t]{0.49\textwidth}
\vspace{0pt}
\begin{tcolorbox}[taskpane={Constraint-specification error}, fonttitle=\bfseries\fontsize{9.6}{11.5}\selectfont]
\textit{Example:} \emph{adulyasak2015} / Claude Opus~4.6
\begin{lstlisting}[style=taskcode]
# x[i,j] is one undirected edge variable
for i in customers:
    for j in customers:
        if i < j:
            model.addConstr(
                u[i] - u[j] + n*x[i,j] <= n-1)
            model.addConstr(
                u[j] - u[i] + n*x[i,j] <= n-1)
\end{lstlisting}
\begin{tcolorbox}[failuremessage]
Gap $77.74\%$ exceeds $10\%$ threshold (obj=$2480.75877$, Gurobi=$1395.75938$).
\end{tcolorbox}
\begin{tcolorbox}[failurecause]
When the undirected edge $x_{ij}=1$, the two MTZ inequalities simultaneously require $u_i<u_j$ and $u_j<u_i$.
\end{tcolorbox}
\end{tcolorbox}
\end{minipage}

\medskip
\noindent
\begin{minipage}[t]{0.49\textwidth}
\vspace{0pt}
\begin{tcolorbox}[taskpane={Interface / schema violation}, fonttitle=\bfseries\fontsize{9.6}{11.5}\selectfont]
\textit{Example:} \emph{barnhart2000} / DeepSeek-R1
\begin{lstlisting}[style=taskcode]
if y[k_id].X > 0.5:
    sol_comm["rejected"] = True
    sol_comm["path_arcs"] = [{
        "from": origin,
        "to": dest,
        "arc_id": -k_id
    }]
\end{lstlisting}
\begin{tcolorbox}[failuremessage]
Commodity 3 references non-existent \texttt{arc\_id -3}; Commodity 4 references non-existent \texttt{arc\_id -4}.
\end{tcolorbox}
\begin{tcolorbox}[failurecause]
The writer invents negative arc IDs for rejected commodities, although these IDs do not exist in the instance network; rejection is already represented by the \texttt{rejected} field.
\end{tcolorbox}
\end{tcolorbox}
\end{minipage}\hfill
\begin{minipage}[t]{0.49\textwidth}
\vspace{0pt}
\begin{tcolorbox}[taskpane={Implementation error}, fonttitle=\bfseries\fontsize{9.6}{11.5}\selectfont]
\textit{Example:} \emph{alfieri2020} / Gemini~3.1~Pro
\begin{lstlisting}[style=taskcode]
f = 0
for v in range(n):
    if conflicts[v] > 0: f += 1
f //= 2                    # conflicting vertices
...
f += best_delta            # conflicting edges
...
if f == 0:
    return color_of         # no revalidation
\end{lstlisting}
\begin{tcolorbox}[failuremessage]
Adjacent vertices 39 and 45 share color 10 ($x_{39,10}+x_{45,10}=2.0>z_{10}=1.0$).
\end{tcolorbox}
\begin{tcolorbox}[failurecause]
The TabuCol state variable changes meaning during updates, so an invalid coloring can be accepted as conflict-free.
\end{tcolorbox}
\end{tcolorbox}
\end{minipage}

\medskip
\noindent
\begin{minipage}[t]{0.49\textwidth}
\vspace{0pt}
\begin{tcolorbox}[taskpane={Heuristic-search failure}, fonttitle=\bfseries\fontsize{9.6}{11.5}\selectfont]
\textit{Example:} \emph{vidal2011} / DeepSeek-R1
\begin{lstlisting}[style=taskcode]
while num_routes > m:
    best_saving = -10**9
    # Search feasible, positive-saving merges
    ...
    if best_saving == -10**9:
        break  # fleet limit may remain unmet
    # Merge the selected pair; update num_routes
for vehicle_index, route in enumerate(routes):
    routes_output.append({
        'vehicle': vehicle_index + 1, ...})
\end{lstlisting}
\begin{tcolorbox}[failuremessage]
Tiny domain violations: routes 5 and 8 use vehicle 3, although valid vehicle IDs are 1--2.
\end{tcolorbox}
\begin{tcolorbox}[failurecause]
The greedy merge search recognizes the fleet limit but stops before satisfying it. Without reassignment, alternate merges, or repair, it emits the remaining infeasible routes.
\end{tcolorbox}
\end{tcolorbox}
\end{minipage}

\subsection{One-shot generation success and failure cases}
\label{app:case-studies-failures}
\begingroup
\color{black}
\tcbset{
  casefailure/.style={
    enhanced, sharp corners, breakable,
    colframe=red!60!black, boxrule=0.5pt,
    colback=red!4, colbacktitle=red!12, coltitle=black,
    fonttitle=\bfseries\scriptsize, fontupper=\scriptsize\color{black},
    titlerule=0pt, left=3pt, right=3pt, top=2pt, bottom=2pt, boxsep=1pt,
  },
  casesuccess/.style={
    enhanced, sharp corners, breakable,
    colframe=green!50!black, boxrule=0.5pt,
    colback=green!4, colbacktitle=green!12, coltitle=green!40!black,
    fonttitle=\bfseries\scriptsize, fontupper=\scriptsize\color{black},
    titlerule=0pt, left=3pt, right=3pt, top=2pt, bottom=2pt, boxsep=1pt,
  },
}
We illustrate correctness, solution quality, and efficiency through three paired comparisons on canonical FrontierOR tasks. Each example presents the task components and contrasts two one-shot solution strategies. Code panels summarize the relevant steps.

\subsubsection{Task 1: Stochastic Unit Commitment}
\label{app:oneshot-salemi2022}
\paragraph{Task overview.} The task associated with Salemi and Davarnia (2022) schedules generator commitments and scenario-dependent production to meet demand and reserve requirements at minimum expected cost. The released task uses a finite-scenario extension with shared commitment decisions. Feasibility requires generation limits, minimum up/down times, and ramping constraints, including the transition from the initially all-off state.

\noindent\begin{tcolorbox}[taskcard={Task 1 (\emph{salemi2022}): Stochastic Unit Commitment}, fontupper=\color{black}]
\textbf{Problem Description.} \textit{A power system must decide when to turn generators on and off and how much electricity each produces under different demand scenarios. Demand and reserve requirements must be met while respecting operating limits and minimizing expected cost.}
\medskip
\begin{minipage}[t]{0.49\textwidth}\vspace{0pt}
\begin{tcolorbox}[taskpane={Mathematical Formulation}, fontupper=\footnotesize\color{black}]
Commitment $x_{it}$ and startup $y_{it}$ are shared across scenarios; production $p_{it\omega}$ is scenario-dependent:
\[
\begin{aligned}
\min\;&\sum_{i,t}(c_i^f x_{it}+q_{it})
       +\frac{1}{|\Omega|}\sum_{i,t,\omega}c_i^g p_{it\omega}\\
\text{s.t.}\;&\sum_i p_{it\omega}\ge D_{t\omega},\\
&p_{it\omega}-p_{i,t-1,\omega}
 \le RU_i x_{i,t-1}+SU_i y_{it},\\
&x_{i0}=p_{i0\omega}=0,\\
&\text{capacity, reserve, and up/down-time rules.}
\end{aligned}
\]
In particular, $p_{i1\omega}\le SU_i y_{i1}$.
\end{tcolorbox}
\begin{tcolorbox}[taskpane={Feasibility Checker}, fontupper=\footnotesize\color{black}]
Checks commitments, production, operating constraints, and objective consistency. The first-period ramp limit is checked as well as transitions between later periods.
\begin{lstlisting}[style=taskcode]
previous_output = 0
previous_on = 0
for t in periods:
    check_ramp(output[t], previous_output,
               on[t], previous_on, startup[t])
    previous_output, previous_on = output[t], on[t]
\end{lstlisting}
\end{tcolorbox}
\end{minipage}\hfill
\begin{minipage}[t]{0.49\textwidth}\vspace{0pt}
\begin{tcolorbox}[taskpane={Test Instance}]
\begin{lstlisting}[style=taskjson]
"instance": "large_1",
"n_generators": 100,
"n_scenarios": 100,
"T": 24
\end{lstlisting}
\end{tcolorbox}
\begin{tcolorbox}[taskpane={Solver Code}]
\begin{lstlisting}[style=taskcode]
model = build_commitment_model(instance)
add_shared_commitment_decisions(model)
add_scenario_production_constraints(model)
add_ramping_from_initial_off_state(model)
model.optimize()
return extract_commitment_and_production(model)
\end{lstlisting}
\end{tcolorbox}
\begin{tcolorbox}[taskpane={Reference Solution}]
\begin{lstlisting}[style=taskjson]
"instance": "large_1",
"objective_value": 5561886.6683,
"status": "time limit with incumbent"
\end{lstlisting}
\end{tcolorbox}
\end{minipage}
\end{tcolorbox}

\paragraph{One-shot failure and success.} Case~\ref{case:salemi2022} contrasts \textbf{omitting an initial-condition constraint with enforcing it explicitly}. Qwen3.8-Max adds ramping constraints only between successive modeled periods, leaving the initial startup unrestricted. Its tiny-instance solution produces $166.88$ units where the startup limit is $140$, and is rejected. Claude Fable 5 explicitly constrains the first period and passes the tiny gate. This comparison concerns correctness, rather than large-instance solution quality or speed.

\refstepcounter{casestudy}\label{case:salemi2022}
\noindent\begin{tcolorbox}[taskcard={Case~\thecasestudy: omitted initial condition (Failure) vs.\ explicit startup constraint (Success)}, fontupper=\color{black}]
\begin{minipage}[t]{0.49\textwidth}\vspace{0pt}
\begin{tcolorbox}[taskpane={(a) Qwen3.8-Max \textnormal{--- \textbf{startup ramp violated}}}]
\begin{lstlisting}[style=taskcode]
for t in periods:
    if t > 0:
        add_ramp_up(p[t], p[t-1], on[t-1], start[t])
        add_ramp_down(p[t-1], p[t], on[t], stop[t])
\end{lstlisting}
\end{tcolorbox}
\begin{tcolorbox}[casefailure, title={[!]\ Failure root cause}]
The first period is excluded, allowing production above the startup ramp limit from the initially all-off state.
\end{tcolorbox}
\end{minipage}\hfill
\begin{minipage}[t]{0.49\textwidth}\vspace{0pt}
\begin{tcolorbox}[taskpane={(b) Claude Fable 5 \textnormal{\textcolor{green!40!black}{--- \textbf{feasible tiny solution}}}}]
\begin{lstlisting}[style=taskcode]
for t in periods:
    if t == 0:
        model.addConstr(p[t] <= SU * start[t])
    else:
        add_ramp_up(p[t], p[t-1], on[t-1], start[t])
\end{lstlisting}
\end{tcolorbox}
\begin{tcolorbox}[casesuccess, title={[+]\ Why it works}]
The initial transition is modeled explicitly; the returned tiny-instance solution satisfies the checker.
\end{tcolorbox}
\end{minipage}
\end{tcolorbox}

\subsubsection{Task 2: Orienteering}
\label{app:oneshot-fischetti1998}
\paragraph{Task overview.} The task associated with Fischetti, Salazar Gonz\'alez, and Toth (1998) selects a closed tour through a subset of prize-bearing locations. It maximizes collected prize subject to a travel-time budget. Selecting profitable locations and arranging their visit order are coupled decisions; the five large instances each contain $500$ nodes.

\noindent\begin{tcolorbox}[taskcard={Task 2 (\emph{fischetti1998}): Orienteering}, fontupper=\color{black}]
\textbf{Problem Description.} \textit{A vehicle starts and ends at a depot and collects a prize at each visited location. Choose which locations to visit and in what order to maximize the total prize without exceeding the travel-time budget.}
\par\medskip
\noindent\begin{minipage}[t]{0.49\textwidth}\vspace{0pt}
\begin{tcolorbox}[taskpane={Mathematical Formulation}, fontupper=\footnotesize\color{black}]
Binary $x_e$ selects edges and $y_v$ selects visits:
\[
\begin{aligned}
\max\;&\sum_v p_v y_v\\
\text{s.t.}\;&\sum_e t_e x_e\le t_0,\\
&\sum_{e\in\delta(v)}x_e=2y_v,\quad y_1=1,\\
&\text{connectivity constraints},\\
&x_e,y_v\in\{0,1\}.
\end{aligned}
\]
\end{tcolorbox}
\begin{tcolorbox}[taskpane={Feasibility Checker}, fontupper=\footnotesize\color{black}]
Checks the closed depot tour, distinct visited locations, connectivity, travel-time budget, and collected prize.
\begin{lstlisting}[style=taskcode]
check_closed_connected_tour(solution)
check_travel_time(solution, limit=t0)
check_reported_prize(solution, prizes)
\end{lstlisting}
\end{tcolorbox}
\end{minipage}\hfill
\begin{minipage}[t]{0.49\textwidth}\vspace{0pt}
\begin{tcolorbox}[taskpane={Test Instance}]
\begin{lstlisting}[style=taskjson]
"instance": "large_1",
"n": 500,
"depot": 1,
"t0": 242
\end{lstlisting}
\end{tcolorbox}
\begin{tcolorbox}[taskpane={Solver Code}]
\begin{lstlisting}[style=taskcode]
model = build_orienteering_model(instance)
add_degree_and_travel_budget_constraints(model)
enforce_tour_connectivity(model)
model.optimize()
return extract_tour(model)
\end{lstlisting}
\end{tcolorbox}
\begin{tcolorbox}[taskpane={Reference Solution}]
\begin{lstlisting}[style=taskjson]
"instance": "large_1",
"objective_value": 17388,
"status": "optimal"
\end{lstlisting}
\end{tcolorbox}
\end{minipage}
\end{tcolorbox}

\paragraph{One-shot failure and success.} Case~\ref{case:fischetti1998} illustrates \textbf{different solution quality at comparable runtime}. Both models return feasible tours on all five large instances and use approximately one hour per instance. Claude Opus 4.6 relies on constructive and local-search heuristics at this scale, leaving prize gaps of $6.83\%$--$20.43\%$. Claude Fable 5 combines a heuristic starting tour with a warm-started MIP and dynamically added connectivity constraints, reaching gaps below $0.01\%$ on every instance. SQ is consequently $0/5$ versus $5/5$.

\refstepcounter{casestudy}\label{case:fischetti1998}
\noindent\begin{tcolorbox}[taskcard={Case~\thecasestudy: heuristic-only search vs.\ hybrid refinement at comparable runtime}, fontupper=\color{black}]
\begin{minipage}[t]{0.49\textwidth}\vspace{0pt}
\begin{tcolorbox}[taskpane={(a) Claude Opus 4.6 \textnormal{--- \textbf{gap $6.83\%$--$20.43\%$}}}]
\begin{lstlisting}[style=taskcode]
if n <= 60:
    solve_exact_model()
else:
    tour = greedy_construct()
    while time_remaining():
        tour = perturb_and_improve(tour)
return best_feasible_tour
\end{lstlisting}
\end{tcolorbox}
\begin{tcolorbox}[casefailure, title={[!]\ Quality limitation}]
At $500$ nodes, the program uses only heuristic search. Feasibility is preserved, but the final prize remains below the reference despite using the full budget.
\end{tcolorbox}
\end{minipage}\hfill
\begin{minipage}[t]{0.49\textwidth}\vspace{0pt}
\begin{tcolorbox}[taskpane={(b) Claude Fable 5 \textnormal{\textcolor{green!40!black}{--- \textbf{gap below $0.01\%$}}}}]
\begin{lstlisting}[style=taskcode]
tour = construct_and_improve()
if n <= 500:
    model = build_edge_and_visit_model()
    warm_start(model, tour)
    model.optimize(connectivity_callback)
return best_feasible_tour
\end{lstlisting}
\end{tcolorbox}
\begin{tcolorbox}[casesuccess, title={[+]\ Why it works}]
Heuristic initialization is combined with MIP refinement at the evaluated scale, yielding near-reference prize values within the same time allowance.
\end{tcolorbox}
\end{minipage}
\end{tcolorbox}

\subsubsection{Task 3: Bin Packing}
\label{app:oneshot-brandao2016}
\paragraph{Task overview.} The task associated with Brand\~ao and Pedroso (2016) packs items into identical capacity-limited bins while minimizing the number of bins. The Gurobi reference uses a compressed arc-flow formulation. We focus on \texttt{large\_2}, where both compared programs reach the reference value of $53$ bins but differ substantially in runtime.

\noindent\begin{tcolorbox}[taskcard={Task 3 (\emph{brandao2016}): Bin Packing}, fontupper=\color{black}]
\textbf{Problem Description.} \textit{Items have given sizes and required quantities. Pack all required items into identical bins without exceeding any bin's capacity, using as few bins as possible.}
\medskip
\begin{minipage}[t]{0.49\textwidth}\vspace{0pt}
\begin{tcolorbox}[taskpane={Mathematical Formulation}, fontupper=\footnotesize\color{black}]
Each source-to-target path in the arc-flow graph represents a feasible bin pattern. Integer arc flows $f_a$ and total flow $z$ satisfy:
\[
\begin{aligned}
\min\;&z\\
\text{s.t.}\;&\text{flow conservation with total flow }z,\\
&\sum_{a:\,\mathrm{type}(a)=i}f_a\ge b_i,\quad\forall i,\\
&f_a\in\mathbb{Z}_{\ge0}.
\end{aligned}
\]
Selected item types use demand equalities.
\end{tcolorbox}
\begin{tcolorbox}[taskpane={Feasibility Checker}, fontupper=\footnotesize\color{black}]
Checks item identifiers and demand coverage, bin capacities, and consistency of the reported bin count.
\begin{lstlisting}[style=taskcode]
check_item_demands(bins, instance)
for bin in bins:
    check_capacity(bin, W)
check_reported_bin_count(solution, bins)
\end{lstlisting}
\end{tcolorbox}
\end{minipage}\hfill
\begin{minipage}[t]{0.49\textwidth}\vspace{0pt}
\begin{tcolorbox}[taskpane={Test Instance}]
\begin{lstlisting}[style=taskjson]
"instance": "large_2",
"item_types": 131,
"total_items": 132,
"bin_capacity": 13333
\end{lstlisting}
\end{tcolorbox}
\begin{tcolorbox}[taskpane={Solver Code}]
\begin{lstlisting}[style=taskcode]
graph = build_compressed_arc_flow_graph(instance)
model = build_integer_flow_model(graph)
add_item_demand_constraints(model)
model.optimize()
return decompose_flow_into_bins(model)
\end{lstlisting}
\end{tcolorbox}
\begin{tcolorbox}[taskpane={Reference Solution}]
\begin{lstlisting}[style=taskjson]
"instance": "large_2",
"objective_value": 53,
"num_bins": 53,
"status": "optimal"
\end{lstlisting}
\end{tcolorbox}
\end{minipage}
\end{tcolorbox}

\paragraph{One-shot failure and success.} Case~\ref{case:brandao2016} contrasts \textbf{equal solution quality with different formulation choices and solving efficiency}. Both programs begin with $54$ bins and reach the reference value of $53$. GPT-5.3-Codex uses item-type-by-bin assignment variables and takes $306.08$ seconds, exceeding the reference runtime of $161.55$ seconds. GPT-5.6-Sol instead uses packing patterns and column generation, returning the same quality in $6.58$ seconds.

\refstepcounter{casestudy}\label{case:brandao2016}
\noindent\begin{tcolorbox}[taskcard={Case~\thecasestudy: bin-indexed MIP vs.\ pattern-based optimization}, fontupper=\color{black}]
\begin{minipage}[t]{0.49\textwidth}\vspace{0pt}
\begin{tcolorbox}[taskpane={(a) GPT-5.3-Codex \textnormal{--- \textbf{$306.08$ s, slower than reference}}}]
\begin{lstlisting}[style=taskcode]
bins = decreasing_packing(items)
model = build_type_by_bin_MIP(bins)
add_demand_and_capacity_constraints(model)
warm_start(model, bins)
model.optimize()
return extract_bins(model)
\end{lstlisting}
\end{tcolorbox}
\begin{tcolorbox}[casefailure, title={[!]\ Efficiency limitation}]
The bin-indexed formulation reaches $53$ bins but takes longer than the reference. Equivalent bins create alternative assignments that the formulation must distinguish.
\end{tcolorbox}
\end{minipage}\hfill
\begin{minipage}[t]{0.49\textwidth}\vspace{0pt}
\begin{tcolorbox}[taskpane={(b) GPT-5.6-Sol \textnormal{\textcolor{green!40!black}{--- \textbf{$6.58$ s, faster than reference}}}}]
\begin{lstlisting}[style=taskcode]
patterns = initial_packing_patterns(items)
while improving_patterns_exist():
    duals = solve_pattern_master(patterns)
    patterns += price_knapsack_patterns(duals)
return solve_integer_pattern_master(patterns)
\end{lstlisting}
\end{tcolorbox}
\begin{tcolorbox}[casesuccess, title={[+]\ Why it works}]
The pattern-based approach optimizes combinations of items within bins and reaches the same bin count substantially faster. Both programs use Gurobi; the distinction is their formulation and solution strategy.
\end{tcolorbox}
\end{minipage}
\end{tcolorbox}
\endgroup

\subsection{Test-time Self-evolution: Improvements, Regressions, and Directions}
\label{app:case-studies}

We illustrate the outcomes in Figure~\ref{fig:self-evolve-task-improvement} with three cases per framework. Each arrow reports the starter's QTE followed by the selected program's QTE, averaged over the same four held-out Test instances. These examples distinguish correctness recovery, quality improvements, runtime failures, and unchanged thresholded scores.

\begin{tcolorbox}[enhanced,breakable,colback=white,colframe=FrontierORHighlight!70!black,
  colbacktitle=FrontierORHighlight,coltitle=black,fonttitle=\bfseries,
  title={OpenEvolve: correctness recovery versus runtime robustness},
  boxrule=0.5pt,left=5pt,right=5pt,top=4pt,bottom=4pt]
\small
\begin{tabularx}{\linewidth}{@{}p{.13\linewidth}p{.18\linewidth}p{.13\linewidth}>{\raggedright\arraybackslash}X@{}}
\textbf{Outcome} & \textbf{Task} & \textbf{Test QTE} & \textbf{Observed change} \\
\midrule
Improved & \emph{eiselt2009} & $0\to1$ & The starter fails objective-consistency checks. The selected hub-search program recomputes its final objective and passes all four Test instances, using cached utility evaluations and budgeted search. \\[4pt]
Regressed & \emph{delage2022} & $1\to0$ & The cutting-pattern program passes the Dev instance but exceeds the one-hour limit on every Test instance. Its Dev success does not transfer to runtime robustness. \\[4pt]
Unchanged & \emph{alfieri2020} & $1\to1$ & Both coloring programs satisfy QTE on all four instances. Continuous objective values change in both directions, but remain within the passing region. \\
\end{tabularx}
\end{tcolorbox}

\begin{tcolorbox}[enhanced,breakable,colback=white,colframe=FrontierORHighlight!70!black,
  colbacktitle=FrontierORHighlight,coltitle=black,fonttitle=\bfseries,
  title={EoH: stronger search need not transfer uniformly},
  boxrule=0.5pt,left=5pt,right=5pt,top=4pt,bottom=4pt]
\small
\begin{tabularx}{\linewidth}{@{}p{.13\linewidth}p{.18\linewidth}p{.13\linewidth}>{\raggedright\arraybackslash}X@{}}
\textbf{Outcome} & \textbf{Task} & \textbf{Test QTE} & \textbf{Observed change} \\
\midrule
Improved & \emph{bierwirth2017} & $0\to1$ & The selected scheduling program combines dispatching, sequence local search, and MIP refinement. All four Test gaps change from $8.4$--$13.5\%$ to better-than-reference solutions. \\[4pt]
Regressed & \emph{martin-iradi2024} & $1\to0.25$ & The berth-scheduling heuristic reduces the Dev gap from $9.5\%$ to $2.5\%$, yet three Test gaps rise above $1\%$. All four Test outputs remain feasible. \\[4pt]
Unchanged & \emph{desaulniers2010} & $0.25\to0.25$ & The selected routing program uses destroy--repair search capped at 20 seconds. It remains feasible on all four Test instances, but still exceeds the quality threshold on three. \\
\end{tabularx}
\end{tcolorbox}

\begin{tcolorbox}[enhanced,breakable,colback=white,colframe=FrontierORHighlight!70!black,
  colbacktitle=FrontierORHighlight,coltitle=black,fonttitle=\bfseries,
  title={CORAL: feasibility recovery and gains below the threshold},
  boxrule=0.5pt,left=5pt,right=5pt,top=4pt,bottom=4pt]
\small
\begin{tabularx}{\linewidth}{@{}p{.13\linewidth}p{.18\linewidth}p{.13\linewidth}>{\raggedright\arraybackslash}X@{}}
\textbf{Outcome} & \textbf{Task} & \textbf{Test QTE} & \textbf{Observed change} \\
\midrule
Improved & \emph{adulyasak2015} & $0\to1$ & The starter fails scenario-variable and balance checks. The selected production--routing program supplies a complete scenario-indexed recourse solution; all four Test instances pass, with gaps at most $0.62\%$. \\[4pt]
Regressed & \emph{watermeyer2020} & $0.75\to0.25$ & Dev quality improves, but Test results include a resource-capacity violation and two feasible outputs above the quality threshold. The selected MIP program includes a fallback that enforces temporal, not full resource, feasibility. \\[4pt]
Unchanged & \emph{desaulniers2014} & $0\to0$ & The inventory--routing program combines delivery construction with route relocation, swaps, and 2-opt. Mean Test gap falls from $34.3\%$ to $7.0\%$, but no instance reaches the $1\%$ threshold. \\
\end{tabularx}
\end{tcolorbox}

\paragraph{Patterns and recommended directions.}
\textbf{(1) Correctness before search:} restoring objective consistency and complete feasible outputs can unlock gains that additional search alone cannot.
\textbf{(2) Robust selection:} Dev improvements can coexist with Test timeouts, quality loss, or constraint violations; broader development checks and validated feasible fallbacks are promising safeguards.

\newpage
\section*{NeurIPS Paper Checklist}

\begin{enumerate}

\item {\bf Claims}
    \item[] Question: Do the main claims made in the abstract and introduction accurately reflect the paper's contributions and scope?
    \item[] Answer: \answerYes{}
    \item[] Justification: The abstract and introduction state that we introduce \emph{FrontierOR}, evaluate twelve LLMs under one-shot generation, and study three self-evolutionary frameworks; the experimental sections verify the headline numbers and restrict the scope to the evaluated models, frameworks, and 180 tasks.

\item {\bf Limitations}
    \item[] Question: Does the paper discuss the limitations of the work performed by the authors?
    \item[] Answer: \answerYes{}
    \item[] Justification: Section~\ref{sec:conclusion} discusses the coverage of the source literature, the scope of the evaluated models and computational settings, the use of fixed instances, and the limits of final-output metrics for assessing anytime performance.

\item {\bf Theory assumptions and proofs}
    \item[] Question: For each theoretical result, does the paper provide the full set of assumptions and a complete (and correct) proof?
    \item[] Answer: \answerNA{}
    \item[] Justification: This work introduces an empirical benchmark and reports experimental results; it contains no formal theorems, lemmas, or proofs.

    \item {\bf Experimental result reproducibility}
    \item[] Question: Does the paper fully disclose all the information needed to reproduce the main experimental results of the paper to the extent that it affects the main claims and/or conclusions of the paper (regardless of whether the code and data are provided or not)?
    \item[] Answer: \answerYes{}
    \item[] Justification: Section~\ref{sec:experimental-setup} describes the model identifiers, task inputs, containerized execution, metrics, and one-shot and test-time evolution protocols. The public benchmark and evaluation repository provide the task files, prompts, scoring code, and run instructions.

\item {\bf Open access to data and code}
    \item[] Question: Does the paper provide open access to the data and code, with sufficient instructions to faithfully reproduce the main experimental results, as described in supplemental material?
    \item[] Answer: \answerYes{}
    \item[] Justification: The benchmark data are available at \url{https://huggingface.co/datasets/frontieror/FrontierOR}, and the evaluation harness is available at \url{https://github.com/Minw913/FrontierOR}. The repository README documents setup, one-shot evaluation, test-time evolution, and computation of the main metrics.
    \item[] Guidelines:
    \begin{itemize}
        \item The answer NA means that paper does not include experiments requiring code.
        \item Please see the NeurIPS code and data submission guidelines (\url{https://nips.cc/public/guides/CodeSubmissionPolicy}) for more details.
        \item While we encourage the release of code and data, we understand that this might not be possible, so "No" is an acceptable answer. Papers cannot be rejected simply for not including code, unless this is central to the contribution (e.g., for a new open-source benchmark).
        \item The instructions should contain the exact command and environment needed to run to reproduce the results. See the NeurIPS code and data submission guidelines (\url{https://nips.cc/public/guides/CodeSubmissionPolicy}) for more details.
        \item The authors should provide instructions on data access and preparation, including how to access the raw data, preprocessed data, intermediate data, and generated data, etc.
        \item The authors should provide scripts to reproduce all experimental results for the new proposed method and baselines. If only a subset of experiments are reproducible, they should state which ones are omitted from the script and why.
        \item At submission time, to preserve anonymity, the authors should release anonymized versions (if applicable).
        \item Providing as much information as possible in supplemental material (appended to the paper) is recommended, but including URLs to data and code is permitted.
    \end{itemize}

\item {\bf Experimental setting/details}
    \item[] Question: Does the paper specify all the training and test details (e.g., data splits, hyperparameters, how they were chosen, type of optimizer) necessary to understand the results?
    \item[] Answer: \answerYes{}
    \item[] Justification: Sections~\ref{subsec:eval-metrics} and~\ref{subsec:eval-protocol} specify the metrics, the tiny-instance checks and runtime-error retries, and the development and held-out test split with a budget of $20$ candidate evaluations per task. The experiments evaluate existing models without training them.

\item {\bf Experiment statistical significance}
    \item[] Question: Does the paper report error bars suitably and correctly defined or other appropriate information about the statistical significance of the experiments?
    \item[] Answer: \answerYes{}
    \item[] Justification: Appendix~\ref{app:statistical-uncertainty} reports task-level bootstrap $95\%$ confidence intervals for held-out test-time evolution and paired QTE differences relative to the starter. These intervals capture variation across tasks for the recorded runs.

\item {\bf Experiments compute resources}
    \item[] Question: For each experiment, does the paper provide sufficient information on the computer resources (type of compute workers, memory, time of execution) needed to reproduce the experiments?
    \item[] Answer: \answerNo{}
    \item[] Justification: Sections~\ref{subsec:computational-environment} and~\ref{subsec:eval-protocol} document the single-core Docker environment, solver settings, runtime limits, and candidate budgets. The paper does not report per-run memory requirements or total measured compute for all experiments.

\item {\bf Code of ethics}
    \item[] Question: Does the research conducted in the paper conform, in every respect, with the NeurIPS Code of Ethics \url{https://neurips.cc/public/EthicsGuidelines}?
    \item[] Answer: \answerYes{}
    \item[] Justification: The research conforms to the NeurIPS Code of Ethics. The benchmark is constructed from publicly available, peer-reviewed operations-research papers; no human-subject data is collected; no personally identifiable information is involved; LLM evaluations are run via standard API endpoints in accordance with each provider's terms of service.

\item {\bf Broader impacts}
    \item[] Question: Does the paper discuss both potential positive societal impacts and negative societal impacts of the work performed?
    \item[] Answer: \answerYes{}
    \item[] Justification: \emph{Positive impacts}: a transparent benchmark for LLM-driven optimization-algorithm design provides a shared yardstick for the OR and ML communities, encouraging reproducible progress on automated solver synthesis for high-stakes decisions in logistics, scheduling, energy, and finance. \emph{Negative impacts}: as with any benchmark for code-generating LLMs, performance numbers may be misinterpreted as real-world reliability claims; we mitigate this by emphasizing the gap to commercial solvers and the structural failure modes (Section~\ref{sec:results-onehot}) so that practitioners do not deploy LLM-generated solvers without independent verification.

\item {\bf Safeguards}
    \item[] Question: Does the paper describe safeguards that have been put in place for responsible release of data or models that have a high risk for misuse (e.g., pre-trained language models, image generators, or scraped datasets)?
    \item[] Answer: \answerNA{}
    \item[] Justification: The released asset is an algorithmic-evaluation benchmark consisting of mathematical optimization problem descriptions, instance generators, Gurobi reference code, and feasibility checkers; it does not contain pretrained models, scraped images, or other high-misuse-risk artefacts.

\item {\bf Licenses for existing assets}
    \item[] Question: Are the creators or original owners of assets (e.g., code, data, models), used in the paper, properly credited and are the license and terms of use explicitly mentioned and properly respected?
    \item[] Answer: \answerYes{}
    \item[] Justification: Every problem in the benchmark cites its source paper, and the natural-language problem descriptions are authored by us based on the cited papers, not copied verbatim. The Gurobi solver is used under an academic license (Gurobi $12$); OpenEvolve, EoH, and CORAL are the official open-source releases of their respective papers and are cited accordingly. All third-party Python libraries (NumPy, SciPy, NetworkX, Pandas, Matplotlib) are used under their permissive licenses (BSD-style or MIT) listed in the supplementary repository.

\item {\bf New assets}
    \item[] Question: Are new assets introduced in the paper well documented and is the documentation provided alongside the assets?
    \item[] Answer: \answerYes{}
    \item[] Justification: Section~\ref{sec:benchmark} describes the benchmark and its task components. The public dataset card and code repository document task files, instance and solution schemas, reference programs, feasibility checkers, and the evaluation harness.

\item {\bf Crowdsourcing and research with human subjects}
    \item[] Question: For crowdsourcing experiments and research with human subjects, does the paper include the full text of instructions given to participants and screenshots, if applicable, as well as details about compensation (if any)?
    \item[] Answer: \answerNA{}
    \item[] Justification: The paper does not involve crowdsourcing or research with human subjects.

\item {\bf Institutional review board (IRB) approvals or equivalent for research with human subjects}
    \item[] Question: Does the paper describe potential risks incurred by study participants, whether such risks were disclosed to the subjects, and whether Institutional Review Board (IRB) approvals (or an equivalent approval/review based on the requirements of your country or institution) were obtained?
    \item[] Answer: \answerNA{}
    \item[] Justification: The paper does not involve research with human subjects, so IRB approval is not applicable.

\item {\bf Declaration of LLM usage}
    \item[] Question: Does the paper describe the usage of LLMs if it is an important, original, or non-standard component of the core methods in this research? Note that if the LLM is used only for writing, editing, or formatting purposes and does \emph{not} impact the core methodology, scientific rigor, or originality of the research, declaration is not required.
    \item[] Answer: \answerYes{}
    \item[] Justification: Section~\ref{subsec:eval-protocol} describes the twelve evaluated LLMs and the three test-time evolution frameworks. Appendix~\ref{app:source-selection} describes the use of Claude Code in constructing benchmark tasks.

\end{enumerate}

\end{document}